\documentclass[journal]{IEEEtran}
\usepackage{cite}
\usepackage{graphicx}
\usepackage{color}
\ifCLASSOPTIONcompsoc
    \usepackage[caption=false,font=normalsize,labelfont=sf,textfont=sf]{subfig}
\else
    \usepackage[caption=false,font=footnotesize]{subfig}
\fi
\usepackage{amsmath}\interdisplaylinepenalty=2500 
\usepackage{amssymb}
\usepackage{mathrsfs}% for \mathscr{}
\usepackage{hyperref}
\usepackage[english]{babel}
\usepackage[utf8]{inputenc}
\usepackage[T1]{fontenc}
\usepackage{multirow}
\usepackage{cmap}
\input{glyphtounicode}
\usepackage{tikz}   
\usepackage{pgfplots}
\usepgfplotslibrary{patchplots}
\usetikzlibrary{plotmarks}
\pgfplotsset{compat=1.16}
\usetikzlibrary{external}
\tikzexternalenable
\newlength\fheight
\newlength\fwidth
\usepackage{ifthen}
\newboolean{shouldLoadPlots} 
\setboolean{shouldLoadPlots}{true}   
\newcounter{TikzCnt}
\newcommand{\tikzOnOverleaf}[1]{%
    \ifthenelse{\boolean{shouldLoadPlots}}{%
        \includegraphics{./figure/generated/fuzzySIT-figure\arabic{TikzCnt}.pdf}%   % Load the externalised image built on another machine
        \addtocounter{TikzCnt}{1}
    }{%
        \input{#1}%  % Compile the actual image
    }% 
}%
\newcommand{\rawepstextikz}[3]{% #1: width, #2: file path in the `./figure/` folder, #3: preamble
    \ifthenelse{\boolean{shouldLoadPlots}}{%
        \centering%    
        \includegraphics[width=#1\linewidth]{./figure/generated/fuzzySIT-figure\arabic{TikzCnt}.pdf}%   % Load the externalised image built on another machine
        \addtocounter{TikzCnt}{1}
    }{%
        \begin{tikzpicture}%
            \node at (0,0){%
                #3%
                \def\svgwidth{#1\linewidth}%
                \input{./figure/#2}%
            };%
        \end{tikzpicture}%
    }% 
}%
\newcommand{\epstextikz}[2]{% #1: width, #2: file path in the `./figure/` folder
    \rawepstextikz{#1}{#2}{\centering\footnotesize}
}%
\usepackage{stmaryrd}
\usepackage{trimclip}
\newcommand{\eg}{\emph{e}.\emph{g}.,~} 
\newcommand{\ie}{\emph{i}.\emph{e}.,~}
\newcommand{\onto}[1]{\text{{\fontfamily{lmss}\selectfont {#1}}}}% ontological font
\newcommand*{\prop}[3]{\ensuremath{{({#1},{#2}){:}{#3}}}}% ex, \prop{\gamma_1}{\gamma_2}{\mathbf{r}} -> (γ1,γ2):R
\newcommand*{\class}[2]{\ensuremath{{{#1}{:}{#2}}}}% ex, \class{γ}{PLANE} -> γ:PLANE 
\newcommand*{\deff}[3]{\ensuremath{{#1}\,{#2}.\ensuremath{#3}}}% ex, \deff{\exists}{A}{b} -> ∃A.b
\newcommand*{\fuzzy}[2]{\ensuremath{{\langle{#1},{#2}\rangle}}}% ex, \fuzzy{a}{.8} -> <a,.8>
\newcommand*{\fuzzyprop}[4]{\fuzzy{\prop{#1}{#2}{#3}}{#4}}% ex, \fuzzyprop{h}{x}{y}{right}{0.8} -> <h(x,y):right,0.8>
\newcommand*{\fuzzyclass}[3]{\fuzzy{\class{#1}{#2}}{#3}}% ex, \fuzzyclass{a}{SPHERE]{.8} -> <a:SPHERE,0.9>
\newcommand*{\fuzzydef}[4]{\fuzzy{\deff{#1}{#2}{#3}}{#4}}% ex, \fuzzydef{\exists}{r}{A}{0,8} -> <∃r.A,0.8>
\newcommand*{\reify}[3]{\onto{#1}\hspace*{.1em}\onto{#2}\hspace*{.1em}\onto{#3}}% ex \reify{A}{B}{C} -> A B C
\newcommand*{\reifyy}[2]{\onto{#1}\hspace*{.1em}\onto{#2}}% ex \reify{A}{B} -> A B
\usepackage{algorithmicx} % or algorithmic, do not use algorithm2e for caption issues
\usepackage{algorithm}
\usepackage{algpseudocode}
\makeatletter
\newcommand{\StatexIndent}[1][3]{%
  \setlength\@tempdima{\algorithmicindent}%
  \Statex\hskip\dimexpr#1\@tempdima\relax}
\algdef{S}[IF]{IfNoDo}[1]{\algorithmicif\ #1}%
\makeatother
\algrenewcommand{\algorithmiccomment}[1]{\hfill\(\triangleright\)~#1}
\algnewcommand{\LineComment}[1]{\vspace{.7em}\Statex\begin{center}\(\triangleright\)~\emph{#1}~\(\triangleleft\)\end{center}\vspace{.2em}}
\algnewcommand{\LongComment}[2]{\hfill\(\triangleright\)~\begin{minipage}[t]{#1\linewidth}#2\strut\end{minipage}}
\newdimen{\algindent}\algnewcommand\Xcomment[2]{\Statex\hspace*{#1\algindent}\(\triangleright\) #2\hfill}
\algnewcommand\ForEach[2]{\For{\textbf{each} #1}#2\EndFor}
\algnewcommand\PRIOR{\item[\textbf{Prior Knowledge:}]}
\algnewcommand\INPUT{\item[\textbf{Input:}]}
\algnewcommand\OUTPUT{\item[\textbf{Output:}]}
\algnewcommand\CONST{\item[\textbf{Constants:}]}
\algnewcommand\STATE{\item[\textbf{State:}]}

\usepackage{xcolor}
\usepackage{pict2e}
\newsavebox{\ORCIDlogo}
\savebox{\ORCIDlogo}{%
\setlength{\unitlength}{\dimexpr .8em/256\relax}%
\begin{picture}(256,256)%
  \color[HTML]{A6CE39}\put(128,128){\circle*{256}}%
  \color{white}%
  \put(78.6,199.2){\circle*{20}}%
  \moveto(70.9,176,9)\lineto(86.3,176,9)\lineto(86.3,69.8)\lineto(70.9,69.8)%
  \closepath\fillpath%
  \moveto(108.9,176.9)\lineto(150.5,176.9)%
  \curveto(190.1,176.9)(207.5,148.6)(207.5 ,123.3)%
  \curveto(207.5,95,8)(186,69.7)(150.7,69.7)%
  \lineto(108.9,69.7)%
  \closepath\fillpath%
  \color[HTML]{A6CE39}%
  \moveto(124.3,83.6)\lineto(148.8,83.6)%
  \curveto(183.7,83.6)(191.7,110.1)(191.7,123.3)%
  \curveto(191.7,144.8)(178,163)(148,163)%
  \lineto(124.3,163)%
  \closepath\fillpath%
\end{picture}%
}
\newcommand\orcidicon[1]{%
        \href{https://orcid.org/#1}{\includegraphics[width=2.7mm]{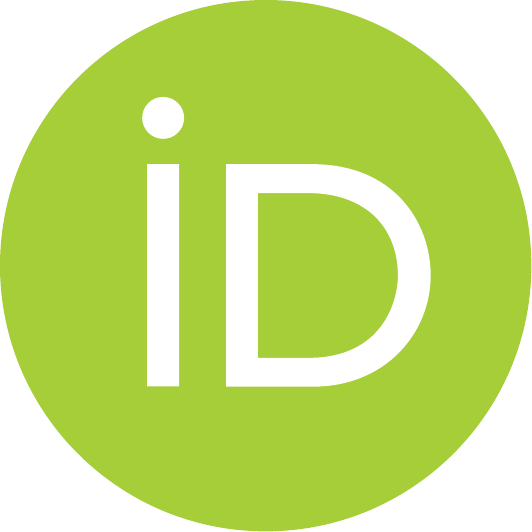}}%
}%

\begin{document}%
\title{Incremental Bootstrapping and Classification of Structured Scenes in a Fuzzy Ontology}%
\ifthenelse{\boolean{shouldLoadPlots}}{%
    % format authors while loading the images, i.e., running on overleaf.
    \author{Luca~Buoncompagni\textsuperscript{\orcidicon{0000-0001-8121-1616}} and Fulvio~Mastrogiovanni\textsuperscript{\orcidicon{0000-0001-5913-1898}}%
        \thanks{L. Buoncompagni and F. Mastrogiovanni
                are affiliated with the Department of Informatics, Bioengineering, Robotics and Systems Engineering, 
                University of Genoa, Via Opera Pia 13, 16145, Genoa, Italy.
                
                Corresponding author: \href{mailto:buon_luca@yahoo.com}{luca.buoncompagni@edu.unige.it}.
                
                We thank Professor Alessandro Saffiotti, affiliated with the University of Örebro in Sweden, to have supported us with the theoretical foundation of the fuzzy logic approach developed in this manuscript.
                
                We also thank Professor Umberto Straccia, affiliated with the National Research Council of Italy, for his hints and software tools to validate our approach.
                
                Manuscript published the 17\textsuperscript{th} April 2024.}
    }%
}{% % do nothing while compiling the actual images (it is required due to a tikz externalize bug)
    \textbf{TODO} recompile while loading images from files to format authors, affiliations, and thanks (see the \emph{shouldLoadPlots} flag on the tex file).
}% 

\newcommand{\keywords}{Incremental Knowledge Bootstrapping, Fuzzy Description Logic, Learn Structured Categories, Human-Robot Interaction.}

\markboth%{\journal\ Class Files,~Vol.~??, No.~??, June~2023}%
         {L. Buoncompagni and F. Mastrogiovanni: I\MakeLowercase{ncremental} B\MakeLowercase{ootstrapping and} C\MakeLowercase{lassification of} S\MakeLowercase{tructured} S\MakeLowercase{cenes in a} F\MakeLowercase{uzzy} O\MakeLowercase{ntology}}{}

% Remember, if you use this you must call \IEEEpubidadjcol in the second column for its text to clear the IEEEpubid mark.

\makeatletter
    \hypersetup{hidelinks,
                pdftitle={\@title},
                pdfauthor={Luca Buoncompagni and Fulvio Mastrogiovanni},
                pdfsubject={A novel approach to create and use fuzzy ontology through human-based demonstrations.},
                pdfkeywords={\keywords},
                pdfproducer={LaTeX},
                pdfcreator={pdfLaTeX}}%
\makeatother
\maketitle

\begin{abstract} % 150-250 words
We foresee robots that bootstrap knowledge representations and use them for classifying relevant situations and making decisions based on future observations.
Particularly for assistive robots, the bootstrapping mechanism might be supervised by humans who should not repeat a training phase several times and should be able to refine the taught representation. 
We consider robots that bootstrap structured representations to classify some intelligible categories. 
Such a structure should be incrementally bootstrapped, \ie without invalidating the identified category models when a new additional category is considered. 
To tackle this scenario, we presented the Scene Identification and Tagging (SIT) algorithm, which bootstraps structured knowledge representation in a crisp OWL-DL ontology. 
Over time, SIT bootstraps a graph representing scenes, sub-scenes and similar scenes. 
Then, SIT can classify new scenes within the bootstrapped graph through logic-based reasoning. 
However, SIT has issues with sensory data because its crisp implementation is not robust to perception noises. 
This paper presents a reformulation of SIT within the fuzzy domain, which exploits a fuzzy DL ontology to overcome the robustness issues. 
By comparing the performances of fuzzy and crisp implementations of SIT, we show that fuzzy SIT is robust, preserves the properties of its crisp formulation, and enhances the bootstrapped representations. 
On the contrary, the fuzzy implementation of SIT leads to less intelligible knowledge representations than the one bootstrapped in the crisp domain.
\end{abstract}


\thispagestyle{empty}

\section{Introduction}

% Why semantic scene representation and issues (i.e., robustness and knowledge bootstrapping)
\noindent
\IEEEPARstart{R}{obots} need representations concerning the environment and the tasks to be performed.
Such representations are usually based on geometrical primitives applied to sensor data.
However, when the robot interacts with humans, it should be aware of common human semantics~\cite{8525527}, and it should also allow users to access and refine knowledge~\cite{VILLAMARGOMEZ2021103763}. 
Crisp \emph{ontologies} have been widely used for semantic representation and human intelligibility based on symbolic structures~\cite{app11104324} but they are affected by robustness issues, since ontological concepts require perfect semantic relations~\cite{4740499}, and this is particularly relevant when noisy sensory data are considered.
Moreover, knowledge \emph{bootstrapping}~\cite{doi:10.1080/09540090600768484}, which brings the gap between semantic concepts and models learning, and knowledge \emph{grounding}~\cite{coradeschi2013short}, which allows symbolic representation to dealing with sensors, are challenging open issues.

% Objectives: incremental learning, intelligible representation, structured classification of similar scenes
We assume a scenario where the robot observes scenes and bootstraps a structured and intelligible representation used to classify similar scenes in the future.
In particular, we focus on ($i$)~an \emph{incremental} learning approach to allow robots bootstrapping their representations while interacting with the environment and humans.
Also, we aim to bootstrap ($ii$)~\emph{intelligible} representations since they would be suitable for human-robot interaction, especially when humans need to supervise the robot.
To enable the classification of \emph{similar} scenes and reasoning algorithms for actuating the robot, we want to bootstrap ($iii$)~\emph{structured} representations involving hierarchies that relates observed scenes to each other.

% Our contributions
With these three objectives, we previously developed the Scene Identification and Tagging algorithm (SIT)~\cite{sit}, which exploits the Ontology Web Language (OWL) based on the Description Logic (DL) formalism~\cite{f2017Introduction}.
Over time, SIT bootstraps scene categories in an ontology from observations, and it classifies future scenes within a graph of scenes previously observed.
As detailed in Section~\ref{sec:SITOverview}, SIT has limitations, some of which are addressed through the fuzzy formalisation of SIT that this paper presents.
As detailed in Section~\ref{sec:contribution}, this paper focuses on an extensions of SIT that is robust to noisy inputs and vague scenes, which could not be handled with the crisp formalisation of SIT previously proposed.

% Verbal ToC
Section~\ref{sec:background} presents the related work, our motivations, and contributions based on previous work.
Section~\ref{sec:DLprimer} introduces the formalism involved in a fuzzy ontology, and Section~\ref{sec:fuzzySIT} proposes the fuzzy extension of SIT.
Section~\ref{sec:result} shows experimental results, which lead to the discussions in Section~\ref{sec:discussions}.

\section{Background}
\label{sec:background}

\subsection{Related Work}

\noindent
% Probabilistic (reinforcement) learning for scene recognition
As surveyed in~\cite{XIE2020107205,app9102110,yuan2021survey}, machine learning techniques (\eg Deep Convolutional Neural Network) have been largely used to classify objects and scenes in support of tasks planning.
These techniques effectively processing raw data, deal with perception noise, and learn generalised classifiers.
The typical outcomes of such classifiers is a synthetic description of an image (or based on multimodal sensors) made by semantic labels.
% Structural learning issues and motivation
Since these labels are usually not structured, they do not provide a sophisticated representation to reason on the relations between elements in the scenes and among different scenes, which is relevant for task planning.
However, there are few work also addressing structured outcomes, \eg~\cite{10.1145/3303766}.
% Catastrophic forgetting issue for incremental learning
Machine learning have also be used in an incremental fashion, but the \emph{catastrophic forgetting}~\cite{FRENCH1999128} is still a challenging open issue, which occur when the overall performance of the classifier decreases due to the introduction of new types of scenes to be classified.
This issue is even more relevant in our scenario, since a human supervisor should bootstrap the robot knowledge for the service of task planning with a few demonstrations.
% Explainable AI for intelligibility
Moreover, machine learning approaches generate models that are usually not intelligible, and this limits the possibility of the human to supervise the robot knowledge.
Although some learning techniques generate intelligible classifiers by design, their performances are topically lower than other approaches generating black-box like classifiers.
Nonetheless, the problem of having explainable models have been addressed either with metrics that help understanding the reasons of a certain outcome, and by coupling machine learning with logic-based symbolic knowledge representation~\cite{10.1145/3561048}.

% Symbolic OWL-based learning with respect to our objectives (i.e., incremental learning, intelligible representation, structured classification of similar scenes)
Symbolic learners have been developed to bootstrap, align or refine the concepts represented in OWL ontologies.
For instance,~\cite{Liao_2019} presents a method to generate ontologies from relational database, and gain the benefits of OWL reasoning.
The Inductive Logic Programming~\cite{cropper2022inductive} involves machine learning mechanisms to induce logic programs by generalising annotated datasets, also within ontological representations, \eg \cite{lisi2013logic}.
Decision Trees and Random Forests have been proposed to learn the representation of symbols in an ontology~\cite{RIZZO2018340}, while approaches based on kernels have been used in~\cite{FANIZZI20121}, on Naive Bayes in \cite{ZHU201530}, on Genetic Programming in~\cite{10.1007/978-3-540-73499-4_66}, and on Reinforcement Learning~\cite{nickles2014interactive}.
In addition,~\cite{Potoniec+2020+195+216} proposes an approach to mind OWL cardinality restrictions.

% Symbolic fuzzy OWL-based learning with respect to our objectives (i.e., incremental learning, intelligible representation, structured classification of similar scenes)
Fuzzy OWL ontologies extends the OWL formalism with fuzzy degrees~\cite{qasim2020comprehensive}, and they require specific reasoners to deal with vague logic axioms, \eg the FuzzyDL reasoner~\cite{4630480}, FIRE~\cite{1705434}, Delorean~\cite{BOBILLO2012258}, GURDL~\cite{haarslev2007optimizing}, GERDS~\cite{habiballa2007resolution}, SoftFacts~\cite{10.1007/978-3-642-05082-4_12}, and LiFR~\cite{10.1007/978-3-319-11955-7_32}.  
Learners has been developed based on the FuzzyDL reasoner involving the identification of data types~\cite{10.1007/978-3-319-91476-3_9}, and concepts inclusions \cite{CARDILLO2022164}.
Fuzzy ontologies have been also developed for spatial reasoning~\cite{bobillospatial}, object detection~\cite{9639989}, and event recognition~\cite{9678479}. 
Learners based on fuzzy logic that do not rely on OWL ontologies have been proposed, \eg Deep Neural Networks have been coupled with fuzzy theories to improve the performances~\cite{9366977}, and intelligibility issues have been also considered, \eg~\cite{8863387}.
In contrast, crisp OWL ontologies have been also used to support probabilistic learners~\cite{10.1093}.

% Crisp and Symbolic learning limitation for our scenario
Although learning approaches that exploits logic-based symbolic formalisms (\eg crisp or fuzzy OWL ontologies) provided some degree of intelligibility and structured outcomes by design, we are not aware of approaches that incrementally bootstrap environmental models from observation over time.
Instead, to bootstrap logic models, symbolic learning approaches rely on annotated datasets processed through a not incremental training phase, \ie if the dataset is augmented, then the training phase should be performed again to avoid catastrophic forgetting.
% Since the training phase is usually expensive and requires parameter tuning, it does not usually rely on a stream of sensory data.

\subsection{Crisp SIT Overview}
\label{sec:SITOverview}

% SIT input interface
\noindent
We previously developed the crisp SIT algorithm~\cite{sit}, which encompass the \emph{encoding}, \emph{learning}, \emph{structuring} and \emph{classification} phases.
As inputs, SIT requires symbolic \emph{facts} $f_i$ that represent the scene through \emph{relations} $\mathbf{r}_z$ between pairs of \emph{elements} ($\gamma_x,\gamma_y$) having relative \emph{types} ($\Gamma_s,\Gamma_h$).
Figure~\ref{fig:facts} shows an example of input facts involving spatial relations $\mathbf{r}$ and objects shapes $\Gamma$, which constitute the SIT \emph{input interface}, \ie the prior knowledge encoded in the ontology.

% SIT encoding phase
In the encoding phase, SIT maps facts into \emph{beliefs}, which represent the currently observed \emph{scene} as an OWL \emph{instance} $\epsilon_t$ in the ontology.
Beliefs are characterised by \emph{cardinalities} $c_{zsh}$ that describe the features of a scene through combinations of $z$-th relations and $sh$-th types, \eg ``2 cups in front of a glass''.

% SIT learning structuring phase
The learning phase occurs when SIT cannot classify an encoded scene with high degree.
In this phase, SIT learns a scene \emph{category} $\Phi_t$, which is stored as a new OWL \emph{concept} in the ontology.
Categories are defined through cardinality \emph{restrictions} $\Omega$, which are stored in the ontology for future classifications based on the the beliefs observed at a certain time instant.
Then, SIT performs the structuring phase, which exploits DL-based reasoning to arrange the categories learned over time in a \emph{memory graph} $M$.
The memory is a hierarchy describing sub-scenes and their similarity through logic implications.
We consider that SIT bootstrap knowledge when categories has been learned and structured in the memory.

% SIT classifying and outcome
Given input facts and the memory bootstrapped in the ontology, SIT exploits DL-based reasoning to perform the classification phase and provide a \emph{classification graph} $M^\star_t$.
The latter is a sub-graph of $M$ that only contains the bootstrapped categories concerning the input scene being classified (\ie the categories having cardinality restrictions \emph{satisfied} by scene beliefs) at a certain time $t$.
Each node of $M^\star_t$ might classify only a part of the scene (\ie a sub-scene), and this fact is measured through a \emph{similarity} value $d^{\Phi_j}_{\epsilon_t}$, which is associate to each $j$-th categories in $M^\star_t$.

% SIT and our contribution, i.e., incremental learning, intelligible representation, structured classification of similar scenes
It is noteworthy that SIT performs one-shot unsupervised learning of scene categories, which can be \emph{incrementally} added in the ontology since a new category do not require to retrain the models of the previous learned categories.
SIT bootstraps a \emph{structured} set of categories that relates scenes having common parts.
Also, SIT provides a classification based on semantic symbols, which are \emph{intelligible} under some degree of intuitiveness.
The classification accuracy and the intelligibility degree that characterise the bootstrapped representation depends on the used input interface.
Such an interface makes SIT general in terms of scenes semantics, but it need to be carefully designed for each application.

% crisp SIT limitations: complexity and over-fitting
The crisp implementation of SIT is computationally complexity since it uses DL reasoning processes for structuring the memory and classifying scenes.
Indeed, the reasoning process is exponentially complex with respect to the number of axioms involved in the ontology~\cite{f2017Introduction}, which depends on the number of symbols encoded in the input interface, and on the number of categories learned in the ontology.
Furthermore, SIT learns a category for each scene that it is not able to classify with a high degree.
This behaviour does not lead to \emph{overfitting}, \ie SIT would not missclassify the scene, but it would provide a more complex classification involving several similar scenes, and this would affect the computation load.
Although Section~\ref{sec:resComplexity} presents the computation complexity of the fuzzy SIT, this paper does not focus on the computation issues specifically.
Nonetheless, Section~\ref{sec:discussions} discusses the approaches we are considering to overcome these issues.

% crisp SIT limitation: robustness 
Another issue of the crisp SIT concerns the \emph{robustness} of $M$ and $M^\star_t$ with respect to noisy and vague inputs facts.
Indeed, learned categories are defined through cardinality restrictions involving crisp threshold-based rules, which might be satisfied (or not) based on beliefs cardinalities.
Thus, small errors in the perception of the beliefs might lead to very different categories bootstrapped in the memory $M$ and, consequently, to different classification outcomes $M^\star_t$.
To overcome this limitation in the crisp domain, we exploited the intelligibility of the bootstrapped categories and the possibility to refine them without the need to retrain all the learned categories.
In~\cite{dialogue_sit}, we presented an architecture where a human verbally interacts with the robot to supervise if cardinality beliefs have been correctly perceived, but this approach demands high effort from the human.
Through the fuzzy extension of SIT presented in this paper, we aim to overcome the robustness issues and make the data supervision mechanism optional.

\subsection{Contribution}
\label{sec:contribution}

% Objectives
\noindent
This paper presents the fuzzy SIT algorithm%
    \footnote{Available at \url{https://github.com/buoncubi/fuzzy_sit}.}%
, which is an extension of the crisp SIT based on a fuzzy ontology subjected to the inferences computed by the FuzzyDL reasoner.
Our objective is to preserve the characteristic of the crisp implementation of SIT, while allowing for a fuzzy representation of beliefs cardinalities and restrictions.
Hence, we aim to improve the robustness of SIT by not considering discontinuities in the identification of satisfied cardinality restrictions, as it occurs in the crisp domain.
In other words, we aim for a smooth transition between bootstrapped scenes representation such to have consistent models even when inputs facts are noisy.

% Results
In a previous work, we preliminary introduced the fuzzy SIT algorithm \cite{fuzzySit_ws} while, in this paper we detail its theoretical foundations, and how it extends its crisp counterpart.
Also, we present experimental results showing that fuzzy SIT improves robustness while having a behaviour consistent with the crisp domain.
% Benefits
Differently from the crisp SIT, fuzzy SIT represents vague scenes with fuzzy relations and elements having more than one \emph{type}.
Such a vague representation allows using input facts that are retrieved from sensors with more sophisticated mechanisms than crisp thresholds.
In addition, fuzzy SIT introduces fuzzy classification and implication degrees, which enrich the bootstrapped memory and the relative classification graphs.

% Drawbacks
Although fuzzy SIT bootstraps more expressive and robust representations than crisp SIT, the latter learns and classifies scenes within a more intelligible representation.
Hence, further work is required for making fuzzy SIT being as much intuitive to human as in the crisp domain.
Another drawback we asses through experiments concerns the scene similarities measure, which is bounded in $[0,1]$ for crisp SIT, while it is bounded in a range that depends on the application setups for fuzzy SIT.

\section{OWL Ontology and Fuzzy DL Primer}
\label{sec:DLprimer}

\noindent
An OWL ontology is a collection of logic \emph{axioms}, which relate \emph{symbols} to represent \emph{things} in the DL formalism.
An OWL \emph{reasoner} infers symbolic \emph{hierarchies} by checking if the axioms are logically consistent.
The reasoner adheres to the \emph{open word} assumption, which assumes that the truthfulness of an axiom is independent of whether or not the related symbols are known.
The DL symbols that occur in ontological axioms are:
    \emph{concepts} (or classes),
    \emph{instances} (or individuals), and
    \emph{roles} (or properties);
as shown in Table~\ref{tab:notation}.

Concepts are structured in a graph of logic implications, while instances are classified into concepts by the mean of roles, which relate pairs of either instances or concepts.
Table~\ref{tab:notation} also summarises the notation we used for generic representations (\eg a concept $\Gamma$), and its name as encoded in an OWL ontologies (\eg the \onto{SPHERE} concept).
The OWL formalism also specifies \emph{concrete concepts} (\eg the data type $\mathbb{R}$), classifying special types of instances (\eg numbers) that can be related to the other instances through \emph{concrete roles}.

An Ontology collects \emph{axioms} $\mathcal{A}_i$, which define symbols and their logic-based semantics.
Table~\ref{tab:notation} details the axioms relevant for this paper, \eg
    ${\onto{SPHERE}\sqsubseteq\onto{SHAPE}}$ (\ie sphere is a type of shape),
    \class{\onto{Sp1}}{\onto{SPHERE}} (\ie \onto{Sp1} is a sphere),
    \prop{\onto{Sp1}}{\onto{0.2}}{\onto{hasRadius}} (\ie \onto{Sp1} has radius 0.2m), and
    $\onto{SPHERE}\,{\triangleq}\,\deff{{=}1}{\onto{hasRadius}}{\onto{METER}}$ (\ie each sphere has exactly one radius).
Also, concepts can be defined as the conjunction (\ie $\sqcap$), or disjunction (\ie $\sqcup$), of other concepts.
Refer to~\cite{f2017Introduction} and~\cite{b2012OWL} for details about crisp OWL-DL axioms.

\begin{table}[t]
    \caption{The Description Logic formalism used in this paper.}
    \label{tab:notation}    
    \centering
    \begin{tabular}{r@{~~}p{.36\textwidth}}\hline\\[-.9em]
            \onto{CONCEPT}   & Capital Greek letter, \eg $\Delta,\Phi,\Gamma$, \emph{etc}.\\
            \onto{role}      & Bold letter, \eg $\mathbf{r},\mathbf{p},\mathbf{q}$, \emph{etc}.\\
            \onto{Instance}  & Greek letter, \eg $\alpha,\beta,\gamma$, \emph{etc}.\\[.05em]
            \hline\\[-.95em]
            $\Delta\sqsubseteq\Phi$   
                             & $\Delta$ is a \emph{sub-concept} of $\Phi$, \ie $\Delta$ \emph{implies} $\Phi$.\\
            \class{\alpha}{\Delta}        
                             & The instance $\alpha$ is \emph{classified} in the $\Delta$ concept.\\
            \prop{\alpha}{\beta}{\mathbf{r}}   
                             & The $\alpha$ instance is \emph{related} with the $\mathbf{r}$ role to the $\beta$ instance.\\%, \eg \prop{\alpha}{\onto{0.3}}{\onto{hasVolume}}.\\ %is the $\mathbf{r}$ role of $\alpha$, \ie $\mathbf{r}$ relates $\beta$ to $\alpha$.\\
            $\Delta\triangleq\deff{\mathscr{R}}{\mathbf{r}}{\Phi}$   
                             & \emph{Define} $\Delta$ to classify instances $\alpha$ with $\mathbf{r}$ spanning in \class{\beta}{\Phi}. 
                               The \emph{cardinality restriction} $\mathscr{R}$ can be: \onto{some}, \onto{only}, \onto{exact $c$} \onto{min $c$} or \onto{max $c$}, 
                               \ie $\{\exists,\,\forall,\,{=}c,\,{\geqslant}c,\,{\leqslant}c\}$, $c\in\mathbb{N}^+$.\\
            $\Delta\triangleq\Phi\sqcap\Pi$
                             & Define $\Delta$ as the \emph{conjunction} of $\Phi$ and $\Omega$.\\
            $\Delta\triangleq\Phi\sqcup\Pi$  
                             & Define $\Delta$ as the \emph{disjunction} of $\Phi$ and $\Omega$.\\\hline
    \end{tabular}
\end{table}

In a fuzzy ontology, some axioms are paired with a fuzzy degree
    $\langle\mathcal{A}_i,p^{\mathcal{A}_i}\rangle$, with  $p^{\mathcal{A}_i}\in\mathbb{R}^{[0,1]}$
\eg %fuzzy axioms could have degree 
    $p^{\Delta\sqsubseteq\Phi}$
    $p^{\class{\alpha}{\Delta}}$, or
    $p^{\prop{\alpha}{\beta}{\mathbf{r}}}$.
Other axioms, such as the one concerning concept definition, their conjunctions and disjunction 
    (\eg ${\Delta\triangleq\deff{\exists}{\mathbf{r}}{\Phi}}$, ${\Phi\sqcap\Pi}$ and ${\Phi\sqcup\Pi}$), 
are paired with a fuzzy \emph{membership} functions, which spans in $[0,1]$; \eg Figure~\ref{fig:shoulder}.

The FuzzyDL reasoner evaluates fuzzy axioms to infer implicit axioms and resolve queries with different logics~\cite{f2016fuzzya}, and we used the Zaden logic.
The classification phase of SIT exploits the implicit axioms concerning the classification of instances into concepts (\eg \class{\alpha}{\Delta}), while the structuring phase relies on the assessments of concepts implications for retrieving a graph of concepts (\eg $\Delta\sqsubseteq\Phi$).

Noteworthy, FuzzyDL cannot reason on \emph{cardinality restriction axioms}
    ${\Delta\triangleq\deff{{\geqslant}c}{\mathbf{r}}{\Phi}}$,
    (\eg ${\Delta\triangleq\deff{{\geqslant}2}{\onto{contain}}{\onto{BALL}}}$).
In a crisp ontology, this type of axioms defines a concept $\Delta$ classifying instances $\alpha$ that are involved in \emph{at least} (\ie ${\geqslant}$) $c\in\mathbb{N}^+$ roles
    \prop{\alpha}{\beta}{\mathbf{r}}, 
with different instances $\beta$ classified in the $\Phi$ concept.
For example, an instance $\alpha\,{\equiv}\,\onto{Supplier}$ would be classified as \class{\alpha}{\Delta} if the ontology encodes at least the axioms:
    \prop{\alpha}{\beta_1}{\onto{contain}}, \class{\beta_1}{\onto{BALL}}, and
    \prop{\alpha}{\beta_2}{\onto{contain}}, \class{\beta_2}{\onto{BALL}}.
    
More generally, FuzzyDL cannot reason on the \onto{exact}, \onto{min} and \onto{max} crisp cardinality restrictions (introduced in Table~\ref{tab:notation}) because the problem of counting fuzzy elements is far from trivial~\cite{f2008Qualifieda}.
Among several proposal~\cite{j2014discussion}, the $\sigma$-count is a popular approach defining the cardinality $c$ as the sum of the fuzzy degrees of all the items in a set.
This approach treats cardinality as a real positive number, \ie $c \in \mathbb{R}^+$, and it computes the \emph{energy} of a set, which is a cardinality measure assuming that the higher the energy value, the more items the fuzzy set should have.
However, the $\sigma$-count cannot discriminate differences between a set having $10$ items with fuzzy degree $0.1$, or one item with degree equal to $1$.

Since SIT is based on cardinality restriction axioms, we exploited the $\sigma$-count approach to compute fuzzy cardinalities $c$, which are stored in the ontology as numbers related to an instance $\epsilon_t$ through a concrete role $\mathbf{r}$, \ie \prop{\epsilon_t}{c}{\mathbf{r}}.
Than, we implement a cardinality restriction through a \emph{left-shoulder} membership function $\Omega$ applied to a concrete concept in the ontology (\ie a fuzzy set, as shown in Figure~\ref{fig:shoulder}), and we define a concept $\Phi\triangleq\deff{\exists}{\mathbf{r}}{\Omega}$.
Hence, $\epsilon_t$ would be classified in $\Phi$, and the relative fuzzy degree $p^{\class{\epsilon_t}{\Phi}}$ would be computed, if the cardinality $c$ \emph{satisfied} the $\Omega$ restriction.
In other words, we classify the instance $\epsilon_t$ based on the fuzzy membership value of $c$ in $\Omega$, as more detailed in Sections~\ref{sec:learning}, \ref{sec:struct}, and~\ref{sec:classify}.
This approach allow defining fuzzy cardinality restrictions that are satisfied based on a continuous membership distribution $\Omega$.
In contrast, crisp cardinality restrictions relies on thresholds, which lead to robustness issues due to discontinuities.

\section{The Fuzzy SIT Algorithm}
\label{sec:fuzzySIT}

\subsection{The Input Interface}
\label{sec:inInterface}

\noindent
SIT defines scenes made of some \emph{elements} $\gamma_j$ that are represented through prior knowledge concerning their \emph{types} $\Gamma$ and \emph{relations} $\mathbf{r}$.
In particular, elements of a scene have fuzzy membership degrees in some of the types
    ${\{\Gamma_1,\ldots,\Gamma_v\}\sqsubseteq\Gamma}$,
and their might be paired by fuzzy relations
    ${\{\mathbf{r}_1,\ldots,\mathbf{r}_w\}\sqsubseteq\mathbf{r}}$.

SIT requires input \emph{facts} 
    ${F_t=\{f_1,\ldots,f_n\}}$, 
which describe a scene at a certain time instant $t$.
Each fact $f_i$ specifies a relation between two elements, \ie
    \prop{\gamma_x}{\gamma_y}{\mathbf{r}_z},
taking also into account the types of the $x$-th and $y$-th elements.
We denote with
    ${\Gamma^x\sqsubseteq\Gamma}$ and 
    ${\Gamma^y\sqsubseteq\Gamma}$
the types of the two elements, \ie 
    ${\class{\gamma_x}{\Gamma_s}, \forall \Gamma_s\sqsubseteq\Gamma^x}$, and
    ${\class{\gamma_y}{\Gamma_h}, \forall \Gamma_h\sqsubseteq\Gamma^y}$.
Thus, given fuzzy degrees 
    ${p_{iz}\triangleq p^{\prop{\gamma_x}{\gamma_y}{\mathbf{r}_z}}}$, 
    ${p_{is}\triangleq p^{\class{\gamma_x}{\Gamma_s}}}$, and 
    ${p_{ih}\triangleq p^{\class{\gamma_y}{\Gamma_h}}}$, 
a fact $f_i$ is defined in the ontology as:
    \begin{equation}\label{eq:in}\begin{aligned}
         f_i =  \bigg\{\fuzzyprop{\gamma_x}{\gamma_y}{\mathbf{r_z}}{\,p_{iz}};\;
               &\Big\{\fuzzyclass{\gamma_x}{\Gamma_s}{\,p_{is}},\; \forall \Gamma_s \sqsubseteq \Gamma^x\Big\};\\
               &\Big\{\fuzzyclass{\gamma_y}{\Gamma_h}{\,p_{ih}},\; \forall \Gamma_h \sqsubseteq \Gamma^y\Big\}\bigg\}.%
    \end{aligned}\end{equation}

\noindent
The \emph{input interface}~\eqref{eq:in} allows using SIT with general-purpose combinations of symbols that imply $\Gamma$ or $\mathbf{r}$, which semantics might be further specified in the ontology.
As an example, Figure~\ref{fig:facts} shows some facts based on prior knowledge involving
    ${\Gamma\sqsupseteq\{\onto{CUP},\onto{GLASS}\}}$, and
    ${\mathbf{r}\sqsupseteq\{\onto{front}\}}$, \ie ${v=2}$ and ${w=1}$.
    
Since an element that occcurs in different relation always has the same type, we could simplify $F_t$ by removing the axioms concerning $\Gamma_s$ and $\Gamma_h$ that are duplicated in different $f_i$.
Hence $F_t$ can be considered as a collection of two sets: one concerns the types of each element \class{\gamma_j}{\Gamma_{j}} (\eg as shown on top of Figure~\ref{fig:facts}), and the other represents the relations between elements \prop{\gamma_x}{\gamma_y}{\mathbf{r}_{z}} (\eg as shown on Figures~\ref{fig:facts1} and \ref{fig:facts2}).

\subsection{Encoding Phase}
\label{sec:encoding}

\noindent
The encoding phase maps $F_t$ into an OWL instance $\epsilon_t$, which represents the \emph{scene} at the current time instant $t$.
Such an instance encodes in the ontology the features of a scene as combinations of \emph{beliefs}, which are reification of facts. % applied as roles to $\epsilon_t$.
The \emph{reify} operator 
    ${\mathcal{R}\big(\mathbf{r}_z, \Gamma_s, \Gamma_h\big)\triangleq\mathbf{r}_{zsh}}$
defines a role through a concatenation of constituent symbols, \eg
    ${\mathbf{r}_{zsh}\equiv\reify{front}{GLASS}{CUP}}$,
which represents that some cups are in front of some glasses.

The crisp implementation of SIT exploits belief encoded in the ontology as
    \prop{\epsilon_t}{\gamma_x}{\mathbf{r}_{zsh}},
and it classifies the scene $\epsilon_t$ based on \emph{minimal cardinality restrictions}, \ie by limiting the minimum number of beliefs for each $zsh$-th combination.
However, as introduced in Section~\ref{sec:DLprimer}, FuzzyDL cannot reason on cardinality restrictions, and we rely the $\sigma$-count approach to represent the cardinalities of the encoded beliefs. 

\begin{figure}[t]
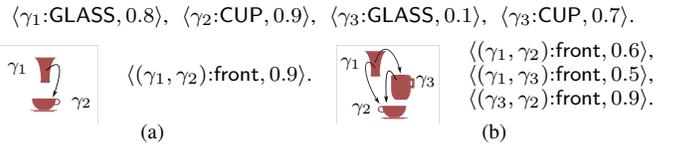

    \centering%
    \footnotesize%
    ~~
    \fuzzyclass{\gamma_1}{\onto{GLASS}}{0.8},\hfill
    \fuzzyclass{\gamma_2}{\onto{CUP}}{0.9},\hfill
    \fuzzyclass{\gamma_3}{\onto{GLASS}}{0.1},\hfill
    \fuzzyclass{\gamma_3}{\onto{CUP}}{0.7}.\hfill
    ~~\\[-.7em]
    \noindent%\hspace{-2ex}
    \subfloat[]{%
        \begin{minipage}{.48\linewidth}%
            \begin{minipage}{.36\linewidth}% 
                \epstextikz{1}{exScene1.eps_tex}%
                \vspace*{-2em}
            \end{minipage}%
            ~~~%
            \begin{minipage}{.4\linewidth}% 
                \centering%
                \fuzzyprop{\gamma_1}{\gamma_2}{\onto{front}}{0.9}.%  
            \end{minipage}%
        \end{minipage}%
        \label{fig:facts1}
    }%
    ~~~%
    \subfloat[]{%
        \begin{minipage}{.48\linewidth}%
            \begin{minipage}{.36\linewidth}% 
                \centering%
                \epstextikz{1}{exScene2.eps_tex}%
                \vspace*{-2em}
            \end{minipage}%
            ~~~%
            \begin{minipage}{.4\linewidth}% 
                \centering%            
                \fuzzyprop{\gamma_1}{\gamma_2}{\onto{front}}{0.6},\\%
                \fuzzyprop{\gamma_1}{\gamma_3}{\onto{front}}{0.5},\\%
                \fuzzyprop{\gamma_3}{\gamma_2}{\onto{front}}{0.9}.%
            \end{minipage}%
        \end{minipage}%
        \label{fig:facts2}        
    }%
    \caption{%
        Examples of input \emph{facts} that represent two \emph{scenes}. 
        The description of the elements types (on top) is in common for both scenes.
    }%
    \label{fig:facts}%
\end{figure}

Hence, fuzzy SIT associates beliefs to the facts in $F_t$ as the conjunction of the $zsh$-th combinations occurring in \eqref{eq:in}, \ie
    \begin{equation}
        \label{eq:cardinalityAxiom}
        \fuzzyprop{\gamma_x}{\gamma_y}{\mathbf{r}_z}{\;p_{iz}} \sqcap \fuzzyclass{\gamma_x}{\Gamma_s}{\;p_{is}} \sqcap \fuzzyclass{\gamma_y}{\Gamma_h}{\;p_{ih}}.
    \end{equation}
Based on the DL Zadeh logic~\cite{f2016fuzzya}, the fuzzy degree of the axiom resulting from \eqref{eq:cardinalityAxiom} is the minimum value among $p_{iz}$, $p_{is}$ and $p_{ih}$.
Thus, the \emph{cardinality} of the $zsh$-th belief computed with the $\sigma$-count approach is
    \begin{equation}
        \label{eq:cardinality}
        c_{zsh} = \sum_{f_i}^{F_t} \min\{p_{iz},\;p_{is},\;p_{ih}\} \in\mathbb{R}^+.
    \end{equation}

\noindent
In the ontology, each $zsh$-th belief is represented with \emph{concrete} crisp roles involving a scene instance $\epsilon_t$, \ie
    \fuzzyprop{\epsilon_t}{c_{zsh}}{\mathbf{r}_{zsh}}{1}.
Beliefs are given by the \emph{encoding function} $\mathcal{E}$, which pairs reified roles and fuzzy cardinalities as
    \begin{equation}
        \label{eq:encoding}
        \mathcal{E}(F_t) \;:\; \{\langle\mathbf{r}_{zsh},c_{zsh}\rangle,\;\;
            \forall \mathbf{r}_z \sqsubseteq \mathbf{r};\; \forall\Gamma_s,\Gamma_h \sqsubseteq \Gamma\}\triangleq B_t.
    \end{equation}

\noindent
For example, the scene in Figure~\ref{fig:facts1} is encoded through one belief, \ie
    $B_t = \{\langle\reify{front}{GLASS}{CUP},0.8\rangle\}$.
Similarly, the set of beliefs $B_t$ concerning Figure~\ref{fig:facts2} is encoded in the ontology as
    \prop{\epsilon_t}{1.2}{\reify{front}{GLASS}{CUP}},
    \prop{\epsilon_t}{0.1}{\reify{front}{GLASS}{GLASS}}, and
    \prop{\epsilon_t}{0.7}{\reify{front}{CUP}{CUP}}.
Noteworthy, not all $zsh$-th combinations always occurs in $B$ due to the open word assumption. % adopted by OWL reasoners.

\begin{figure*}[t]
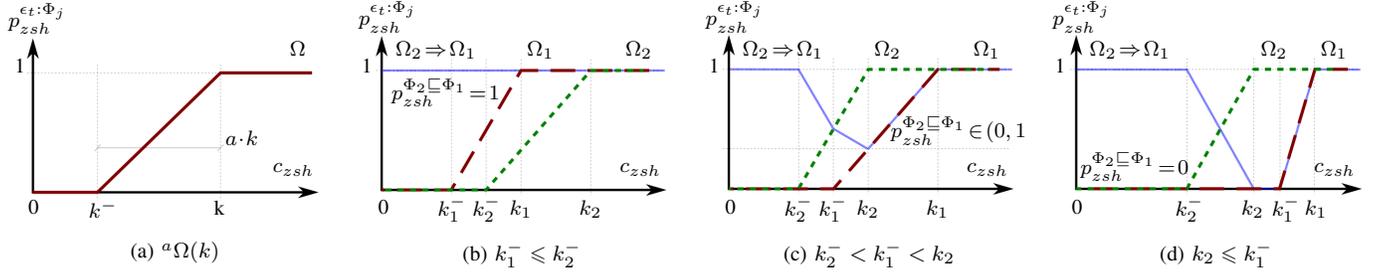
%
    %\vspace*{1em}%
    \footnotesize
    \centering%
    \subfloat[$^a\Omega(k)$]{%
        \epstextikz{.24}{sholder1.eps_tex}%
        \label{fig:shoulder}%
    }%
    ~%\qquad%
    \subfloat[$k_1^- \leqslant k_2^-$]{%
        \epstextikz{.24}{sholder2.eps_tex}%
        \label{fig:restrictionYES}%
    }%
    ~%\qquad%
    \subfloat[$k_2^-<k_1^-<k_2$]{%
        \epstextikz{.24}{sholder3.eps_tex}%
        \label{fig:restrictionFUZZY}%
    }%
    ~%\qquad%
    \subfloat[$k_2 \leqslant k_1^-$]{%
        \epstextikz{.24}{sholder4.eps_tex}%
        \label{fig:restrictionNO}%
    }%
    \caption{%
        Figure (a) shows the membership function $\Omega$, which defines a minimal \emph{cardinality restriction} to $k$.
        Given a \emph{fuzziness} value $a\in[0,1]$ and a cardinality $c_{zsh}\in\mathbb{R}^+$, ${^{a}\Omega}(k)$ specifies the fuzzy degree $p^{\class{\epsilon_t}{\Phi_j}}_{zsh}$ that represents if $c_{zsh}$ satisfies the restriction.  
        The other figures show two cardinality restrictions $\Omega_1$ (in red) and $\Omega_2$ (in green), which are associated to a $zsh$-th beliefs combination of the learned category $\Phi_1$ and $\Phi_2$, respectively.
        Figures (b), (c) and (d) show the implication of $\Omega_1$ and $\Omega_2$ (in blue), which minimum value is the \emph{subsumption degree} $p^{\Phi_1\sqsubseteq\Phi_2}_{zsh}$, and it is used to identify implications between scenes categories, \ie the structure of bootstrapped knowledge into a graph. 
        In particular, (b) shows the case where the minimal cardinality restriction $\Omega_2$ always implies $\Omega_1$ (\ie $\Omega_2$ respects the restriction $\Omega_1$), while (d) shows the case where $\Omega_2$ never implies $\Omega_2$, and (c) show the case where the implication has a fuzzy degree in (0,1).
    }%
    \label{fig:sholders}
\end{figure*}

\subsection{Learning Phase}
\label{sec:learning}

\noindent
The learning phase evaluates the beliefs of the current scene $\epsilon_t$ to generate minimum cardinality restrictions for classifying similar scenes in the future.
Based on the $\sigma$-count approach applied to a fuzzy ontology, a minimum cardinality restriction is a \emph{concrete} concept $\Omega$ that defines a fuzzy set with a \emph{left-shoulder} membership function.
As shown in Figure~\ref{fig:shoulder}, such a function maps a belief cardinality $c_{zsh}{\in}\mathbb{R}^+$ into a fuzzy degree $p^{\class{\epsilon_t}{\Phi_j}}_{zsh}$, which contributes to the classification of future scenes as far as the $zsh$-th combination is concerned.

We parametrize such a membership function as $^{a}\Omega(k)$ where
    $k$ is equal to the $c_{zsh}$ cardinality that was encoded in $\epsilon_t$ at a certain time,
    and $a{\in}[0,1]$ is the \emph{fuzziness} value used to compute $k^- {=} k{\cdot}(1-a)$.
If $a {=} 0$, $\Omega$ becomes a crisp cardinality restriction while, if $a {=} 1$, a non zero degree would be assigned to all $c_{zsh} {>} 0$.
Given $^{a}\Omega(k)$ generated in a previous time instant, and a cardinality value $c_{zsh}$ related to the current scene $\epsilon_t$, we would consider the minimal fuzzy restriction \emph{satisfied} if 
    $c_{zsh} {\geqslant} k$, \ie $p_{zsh} {=} 1$.
Instead, when
    $c_{zsh} {\leqslant} k^-$,
the restriction would be \emph{not satisfied}, \ie $p_{zsh} {=} 0$, while if
    $c_{zsh} {\in} (k^-,k)$
then the restriction will be \emph{fuzzy satisfied}, \ie $p_{zsh} {\in} (0,1)$.

Given encoded beliefs $B_t$ related to $\epsilon_t$ \eqref{eq:encoding}, the \emph{learning function} $\mathcal{L}$ defines a new scene \emph{category} $\Phi_t$ that restricts each $zsh$-th combination occurring in $B_t$ with a relative $\Omega$.
To define the learned categories as nodes of the memory graph, we pose $\Phi_t\sqsubseteq\Phi$, where $\Phi$ is a concept defined in the ontology as prior knowledge to represent the empty scene, \ie $\Phi$ is the root of the memory graph.
Formally, $\mathcal{L}$ allows learning categories $\Phi_t$ as conjunctions of cardinality restrictions, \ie
    $\deff{\exists}{\mathbf{r}_{zsh}}{{^{a}\Omega(c_{zsh})}}$ 
with a fuzzy degree equal to 1; namely,
    \begin{equation}
        \label{eq:learn}
        \mathcal{L}\big(B_t,a\big) \;:\; 
            \fuzzy{\Phi}{\,1} \bigsqcap_{zsh}^{B_t} \fuzzydef{\exists}{\mathbf{r}_{zsh}}{{^{a}\Omega(c_{zsh})}}{1} 
                \triangleq \Phi_t.
    \end{equation}

\noindent
For instance, given the beliefs of the scene shown in Figure~\ref{fig:facts1}, SIT would learn a category $\Phi_t$ with the restriction
    $\deff{\exists}{\reify{front}{GLASS}{CUP}}{{^{a}\Omega(0.8)}}$.
As introduced in Section~\ref{sec:DLprimer}, and coherently with the crisp implementation of SIT, such a restriction indicates that a future scene can be classified in $\Phi_t$ if ``\emph{at least} 0.8 cups are in front of glasses''.
Similarly, the category learned from the scene shown in Figure~\ref{fig:facts2} is
\begin{align}
    &\Phi_t \triangleq \;\Phi \;\sqcap\; 
          \deff{\exists}{\reify{front}{GLASS}{CUP}}{{^{a}\Omega(1.2)}}   \,\sqcap\\
        & \deff{\exists}{\reify{front}{GLASS}{GLASS}}{{^{a}\Omega(0.1)}}   \sqcap
          \deff{\exists}{\reify{front}{CUP}{CUP}}{{^{a}\Omega(0.7)}}.      \notag            
\end{align}
Since FuzzyDL respects the open word assumption, some $zsh$-th combination are unspecified in $\Phi_t$, but they might occur in the beliefs of a future scene classified in the $\Phi_t$ category.

\subsection{Structuring Phase}
\label{sec:struct}

\noindent
The structuring phase exploits the FuzzyDL reasoner to generate a memory graph $M$ having previously learned categories as nodes (\eg $\Phi_1$, $\Phi_2$, \emph{etc.}), while edges are based on fuzzy subsumption axioms.
In particular, a node in the graph $\Phi_2$ that is a sub-scene category of $\Phi_1$ is represented in the ontology with the subsumption axiom
    \fuzzy{\Phi_2\sqsubseteq\Phi_1}{p^{\Phi_2\sqsubseteq\Phi_1}}.
Since we adopted the DL Zadeh logic~\cite{f2016fuzzya}, and since $\mathcal{L}$~\eqref{eq:learn} defines a conjunction of restrictions, the subsumption degree $p^{\Phi_2\sqsubseteq\Phi_1}$ is the minimum $p^{\Phi_2\sqsubseteq\Phi_1}_{zsh}$ value among the $zsh$-th combinations restricted either by $\Phi_1$ and $\Phi_2$.
However, if a $zsh$-th combination occurs in $\Phi_1$ but does not occur in $\Phi_2$, we consider $p^{\Phi_2\sqsubseteq\Phi_1} {=} 0$.
% since all restrictions should be satisfied by the subsuming category.
Instead, if a $zsh$-th combination occurs in $\Phi_2$ but does not occur in $\Phi_1$, FuzzyDL relies on the open word assumption to disregard such a $zsh$-th combination while computing $p^{\Phi_2\sqsubseteq\Phi_1}$.
%, because the subsuming category can always restrict more belief than its parent.
    
For each categories pair $(\Phi_1,\Phi_2)$ restricting the same $zsh$-th combination respectively with $^a\Omega_1(k_1)$ and $^a\Omega_2(k_2)$, FuzzyDL computes the $p^{\Phi_2\sqsubseteq\Phi_1}_{zsh}$ degree as the minimum value of the membership function representing the implication ${\Omega_2\Rightarrow\Omega_1}$.
In particular, the $p^{\Phi_2\sqsubseteq\Phi_1}_{zsh}$ degree is computed based on the three following situations.
    If $k_1^- \leqslant k_2^-$ (Figure~\ref{fig:restrictionYES}) the restriction $\Omega_2$ will always \emph{respects} $\Omega_1$,
\ie $\Phi_2\sqsubseteq\Phi_1$ with $p^{\Phi_2\sqsubseteq\Phi_1}_{zsh}=1$.
% In this case, an instance \class{\epsilon_t}{\Phi_2} is always an instance of $\Phi_1$ with an high degree as well.
%
    When ${k_2^- < k_1^- < k_2}$ (Figure~\ref{fig:restrictionFUZZY}) the restriction $\Omega_2$ \emph{fuzzy respects} $\Omega_1$, 
\ie ${p^{\Phi_2\sqsubseteq\Phi_1}_{zsh}\in(0,1)}$. 
% , and an instance $\epsilon_t$ can be classified in $\Phi_1$ with an higher degree than $\Phi_2$.
%
    When $k_2 < k_1^-$ (Figure~\ref{fig:restrictionNO}), the restriction $\Omega_2$ does \emph{not respect} $\Omega_1$,
thus, $\Phi_2\not\sqsubseteq\Phi_1$, \ie $p^{\Phi_2\sqsubseteq\Phi_1}_{zsh}=0$.
% In other words, an instance $\epsilon_t$ classified in $\Phi_2$ cannot be an instance of $\Phi_1$ with a degree higher than 0.

We map subsumption axioms in a partially ordered set $M$, which is an oriented weighted graph with a root node $\Phi$.
In $M$, the axiom ${\Phi_2\sqsubseteq\Phi_1}$ would be represented as an edge departing from the child $\Phi_2$ and entering in the parent $\Phi_1$ with a weight ${p^{\Phi_2\sqsubseteq\Phi_1}>0}$; \ie an implication arrow, as shown in Figures~\ref{fig:scenariosM1} and \ref{fig:scenariosM2}.
A child node always restricts more beliefs than its parents, \ie the child categorises a more complex scene that contains its parents, which are sub-scenes.

For instance, if the fuzziness value $a$ is such that the restriction learned from the belief
    \prop{\epsilon_2}{1.2}{\reify{front}{GLASS}{CUP}}
satisfies
    \prop{\epsilon_1}{0.8}{\reify{front}{GLASS}{CUP}}
with a degree ${p^{\Phi_2\sqsubseteq\Phi_1}_{121}=0.9}$,
then the scene $\Phi_2$ (shown in Figure~\ref{fig:facts2}) would be a child of $\Phi_1$ (shown in Figure~\ref{fig:facts1}) with subsumption degree that involves a single $zsh$-th combination, \ie
    ${{p^{\Phi_2\sqsubseteq\Phi_1}} = \min\{{p^{\Phi_2\sqsubseteq\Phi_1}_{121}}\}}$.

We formally indicate the DL-based reasoning process to generate the \emph{memory graph} $M$ with the \emph{structuring function} $\mathcal{S}$, which can be expressed in an incremental fashion as
    \begin{equation}
        \label{eq:struct}
        \mathcal{S}\big(M_{t-1}, \Phi_t\big) :
            \big\{\Phi,\Phi_1,\Phi_2,\ldots,\Phi_{t-1},\Phi_t;{\sqsubseteq}\big\} {\triangleq} M_{t}.
    \end{equation}
Given a new category $\Phi_t$, and an initial memory ${M_0 = \{\Phi\}}$, $\mathcal{S}$ updates the edges of $M_{t-1}$ to compute $M_t$.
FuzzyDL performs $\mathcal{S}$, which results are retrieved by quering the subsumption degrees 
    $p^{\Phi_l\sqsubseteq\Phi_j}\; \forall \Phi_l,\Phi_j\sqsubseteq\Phi$ with $l\neq j$.
%To perform such queries, \eqref{eq:learn} introduces the root $\Phi$, which allow retrieving all the categories $\Phi_l$ and $\Phi_j$ learned so far. 
%, \ie all the concepts that subsume $\Phi$; formally, $\forall\Phi_i:\;p^{\Phi_i\sqsubseteq\Phi}>0$.

\subsection{Classification Phase}
\label{sec:classify}

\noindent
In the classification phase, SIT attempts to classify a scene $\epsilon_t$ within some nodes $\Phi_j$ of a memory graph $M_t$ previously structured
A classification \class{\epsilon_t}{\Phi_j} occurs when all the 
   \deff{}{\mathbf{r}_{zsh}}{^a\Omega(k)}
restrictions of $\Phi_j$ are satisfied%
   \footnote{Accordingly with the second paragraph of Section~\ref{sec:learning} and Figure~\ref{fig:shoulder}.}
by the $c_{zsh}$ cardinalities of $\epsilon_t$ encoded in the ontology \eqref{eq:encoding}.
Since $\mathcal{L}$~\eqref{eq:learn} defines conjunctions of cardinality restrictions, the \emph{classification degree} 
    $p^{\class{\epsilon_t}{\Phi_j}}$ 
is the minimum $p^{\class{\epsilon_t}{\Phi_j}}_{zsh}$ degree among all the $zsh$-th combinations restricted by $\Phi_j$.
If $p^{\class{\epsilon_t}{\Phi_j}}=0$, we would consider the scene $\epsilon_t$ as not classified in $\Phi_j$.

Given the structure of the memory graph $M_t$, the scene $\epsilon_t$ would be classified on its sub-graph because each parent of a classified category $\Phi_j$ also are classifications of $\epsilon_t$, \ie sub-scenes.
This sub-graph is named \emph{classification graph} $M^\star_t$ and is given by the \emph{classification function} $\mathcal{C}$, which is computed by FuzzyDL.
The outcomes of $\mathcal{C}$ are retrieved by querying all categories in $M_t$ with a classification degree $p^{\class{\epsilon_t}{\Phi_j}} {>} 0$, \ie 
    \begin{equation}
        \label{eq:classify}
        \mathcal{C}\big(M_t,B_t\big) \;:\;
            \big\{\forall \Phi_j \in M_t\, |\, p^{\class{\epsilon_t}{\Phi_j}}>0;{\sqsubseteq}\big\} \triangleq M^\star_t.
    \end{equation}

\noindent
The crisp implementation of SIT defines the \emph{similarity} value to discriminate categories that are classified based on few restrictions.
Indeed, $M^\star_t$ might involve a category with classification degree $p^\class{\epsilon_t}{\Phi_j}=1$ even if $\epsilon_t$ is more complex than $\Phi_j$, \ie $\epsilon_t$ has higher cardinalities and more beliefs that $\Phi_j$, which classifies only a small sub-scene of $\epsilon_t$.
Analogously, the fuzzy similarity $d^{\Phi_j}_{\epsilon_t}$ can be defined as the ratio between the $zsh$-th restrictions of $\Phi_j$ and the $c_{zsh}$ cardinalities of $\epsilon_t$
    \begin{equation}
        \label{eq:similarity}
        d^{\Phi_j}_{\epsilon_t} \triangleq \frac{d^{\Phi_j}}{d_{\epsilon_t}} = \frac{\sum_{\mathbf{r}_{zsh}}^{\Phi_j}\,k_{zsh}}{\sum_{\mathbf{r}_{zsh}}^{B_t}\,c_{zsh}}.
    \end{equation}

\noindent
In the crisp case, $d^{\Phi_j}_{\epsilon_t}\in[0,1]$ since ${d_{\epsilon_t}{\geqslant}d^{\Phi_j}}$, otherwise, $\Phi_j$ would not be a node of $M^\star_t$ because its restrictions would not be satisfied. 
In the fuzzy domain, the similarity is still bounded but it might be bigger than $1$ when $c_{zsh}$ is within the range defined by the fuzziness parameter $a$.
As shown in Figure~\ref{fig:similExceed} (presented in Section~\ref{sec:result}), the more the similarity exceeds $1$, the lower $p^{\class{\epsilon_t}{\Phi_j}}$ becomes.
Moreover, the fuzzy similarity value  $d^{\Phi_j}_{\epsilon_t}$ is not linear with the number of scene elements but with the number of input facts; consistently with the crisp domain.

For instance, the scene $\epsilon_2$ encoded from Figure~\ref{fig:facts2} would be classified in the category $\Phi_1$ learned from Figure~\ref{fig:facts1} with a degree 
    $p^{\class{\epsilon_2}{\Phi_1}}=1$
because the \reify{front}{GLASS}{CUP} belief of the scene satisfies the restriction of the category.
However, since some of the scene beliefs are not restricted by the classifying category, the similarity value would be 
    $d^1_2=0.8/2$.
      
It is noteworthy that SIT provides an expressive data structure $M^\star_t$. Indeed, $M^\star_t$ inherits from $M_{t-q}$ the cardinality restrictions \eqref{eq:learn} associated to each category classifying a scene $\epsilon_t$, and the subsuming degree $p^{\Phi_l\sqsubseteq\Phi_j}$ between the relative nodes.
In addition, each node of $M^\star_t$ also encodes a classification degree $p^\class{\epsilon_t}{\Phi_j}$ and a similarity value $d^{\Phi_j}_{\epsilon_t}$, which measure how well the restrictions of a category satisfy scene beliefs, and how many beliefs are satisfied by that category, respectively.

\subsection{Complexity Reduction}
\label{sec:simplify}

\noindent
The scene features that SIT represents in the ontology span in a number of reified relation $\mathbf{r}_{zsh}$ equal to $w{\cdot}v^2$, \ie all the combinations between a relation $\mathbf{r}$ and two element types $\Gamma$~\eqref{eq:in}.
Since DL encompass the open word assumption, the number of axioms subjected to reasoning is typically reduced.
However, DL reasoning is exponentially complex with the amount of axioms in the ontology.
Hence, the computation performances are affected by $w$, $v$ and the number of categories in $M$.

To reduce the complexity of SIT, we can change the reification operator presented in Section~\ref{sec:encoding} into
    ${\mathcal{R}\big(\mathbf{r}_z, \Gamma_h\big)\triangleq\mathbf{r}_{zh}}$,
which defines beliefs involving a generic element type, \eg some cups are in front of some \emph{objects}, \ie \reifyy{front}{CUP}.
In this case, a belief that was defined in~\eqref{eq:cardinalityAxiom} becomes,
\begin{equation}
    \label{eq:encodeSimplify} 
    {\bigsqcup_{\forall\Gamma_s\sqsubseteq\Gamma^x}}
        \Big\{ \fuzzyprop{\gamma_x}{\gamma_y}{\mathbf{r}_z}{p_{iz}} \sqcap \fuzzyclass{\gamma_x}{\Gamma_s}{p_{is}} \sqcap \fuzzyclass{\gamma_y}{\Gamma_h}{p_{ih}} \Big\},
\end{equation}
and its cardinality, computed as in~\eqref{eq:cardinality}, becomes
\begin{eqnarray}
    \label{eq:cardinalitySimplified}
    {c_{zh} = \sum_{f_i}^{F} \max_{\forall\Gamma_s\sqsubseteq\Gamma^x}\{\min\{p_{iz},p_{is},p_{ih}\}\big\}}.
\end{eqnarray}
In this way, the number of reified relations $\mathbf{r}_{zh}$ is reduced to $w{\cdot}v$, and all the formalisation presented in Section~\ref{sec:fuzzySIT} remains consistent if the $zsh$-th combination is replaced with the $zh$-th.

% Singolarity and the importance of the input interface
This simplification reduces the amount of scenes features that the algorithm exploits to bootstrap the memory graph, which becomes less intelligible. 
Also, this simplification can lead to \emph{singularities}, which occur when two different scenes are encoded with the same beliefs. 
In general, singularities occur when an input interface does not capture the scene features relevant for the application, but this simplification aggravates such an issue because $\Gamma_s$ is disregarded even if it might refer to a characteristic feature of the scene.

% Inverse relation
Nevertheless, the singularities that this simplification would introduce are avoided when the input interface involves \emph{inverse} relations $\mathbf{r}_z^{-1}$, \eg \onto{front}  and \onto{behind}. 
Hence, this simplification might increase $w$, and $2w{\cdot}v$ combinations occur, in the worst case.
With inverse relations, the information \class{\gamma_x}{\Gamma_s} that this simplification disregards from the fact \fuzzyprop{\gamma_x}{\gamma_y}{\mathbf{r}_z}{p_{iz}}, would be considered in \fuzzyprop{\gamma_y}{\gamma_x}{\mathbf{r}_z^{-1}}{p_{iz}^{-1}}.
If \emph{symmetric} relation are also considered, \ie ${p_{iz}^{-1}=p_{iz}}$, than the beliefs would have balanced relations between inverse pairs of elements.

\begin{algorithm}[t]
    \footnotesize
    \caption{The fuzzy SIT algorithm.}
    \label{alg:algorithm}
    \begin{algorithmic}[1] % The number tells where the line numbering should start
        \PRIOR  $v$ element types $\Gamma$, and $w$ relations $\mathbf{r}$.
        \STATE  a fuzzy ontology $\mathcal{O}$ with the related reasoner.
        \CONST  the membership threshold ${\texttt{th}_m\in[0,1]}$ the similarity threshold ${\texttt{th}_s\geqslant 0}$, and the fuzziness value ${a\in[0,1]}$.
        \INPUT  a set of facts $F_t$ representing the current scene as in~\eqref{eq:in}.
        \OUTPUT the classification graph $M^\star_t$ as defined in~\eqref{eq:classify}.
\LineComment{Encode facts of the current scene into beliefs.}
        \State $d_{\epsilon_t} \gets 0$
\label{ln:econdingStart}
        \State $B \gets \mathcal{E}\big(F\big)$
\Comment{Map the beliefs of the scene computing~\eqref{eq:encoding} based on~\eqref{eq:cardinalitySimplified}.}
        \ForEach{$\langle\mathbf{r}_{zh}, c_{zh}\rangle \in B$}{%
                \State $\mathcal{O}\texttt{.assert}(\fuzzyprop{\epsilon_t}{c_{zh}}{\mathbf{r}_{zh}}{1})$
\Comment{Define the scene $\epsilon_t$ in $\mathcal{O}$.}
                \State $d_{\epsilon_t} \gets d_{\epsilon_t} + c_{zh}$
        }
\label{ln:econdingEnd}
\LineComment{Classify the scene into categories.}
        \State $M_t \gets \mathcal{O}\texttt{.query}(\Phi_j\sqsubseteq\Phi)$ 
\label{ln:classifyStart}
\Comment{Get all sub classes of $\Phi$ in $\mathcal{O}$.}
        \State $M^\star_t \gets \mathcal{C}(M_t,B_t)$ 
\Comment{Reason on $\mathcal{O}$ to query $p^{\class{\epsilon_t}{\Phi_j}}>0$ as in~\eqref{eq:classify}.}
\label{ln:classify}
        \ForEach{$\Phi_j \in M^\star_t$}{
                \State $d^{\Phi_j} \gets \sum_{\mathbf{r}_{zh}}^{\Phi_j}\,k_{zh}$
                \State $M^\star_t\texttt{.setSimilarity}(\Phi_j, d^{\Phi_j}/d_{\epsilon_t})$ 
\Comment{Assign $d^{\Phi_j}_{\epsilon_t}$ to $\Phi_j$~\eqref{eq:similarity}.}
\Xcomment{2}{Line~\ref{ln:classify} also assigns the membership degrees $p^{\class{\epsilon_t}{\Phi_j}}$ to each $\Phi_j$.}
        }
\label{ln:classifyEnd}        
        \LineComment{Eventually, learn and structure a new scene category from beliefs.}
        \If{$\max_{\Phi_j}^{M^\star_t}\{p^{\class{\epsilon_t}{\Phi_j}}\} < \texttt{th}_m$ \textbf{or}
            $\max_{\Phi_j}^{M^\star_t}\{d^{\Phi_j}_{\epsilon_t}\} < \texttt{th}_s$}
\label{ln:ifStatament}
            \State $\Phi_t \gets \mathcal{L}(B,a)$
\label{ln:learn}
\Comment{Define a new scene category as in~\eqref{eq:learn}.}
            \State $M_t \gets \mathcal{S}(M_t, \Phi_t)$ 
\label{ln:struct}
\Comment{Update $\mathcal{O}$ and query $\Phi_t\sqsubseteq\Phi_j$ as in~\eqref{eq:struct}.}
            \State \textbf{compute} Lines~\ref{ln:classify}--\ref{ln:classifyEnd}
\label{ln:reclassify}
\Comment{Update the classification of $\epsilon_t$ in $M^\star_t$.}
        \EndIf
\vspace{.7em}
        \State $\mathcal{O}\texttt{.remove}(\epsilon_t)$
\label{ln:clean}
\Comment{Clean all the beliefs of the scene in $\mathcal{O}$.}
        \State\Return $M^\star_t$
    \end{algorithmic}
\end{algorithm}

\subsection{Implementation}

\noindent
Algorithm~\ref{alg:algorithm} implements the simplification presented in Section~\ref{sec:simplify}, and it provides a classification graph $M^\star_t$ given the facts $F_t$ describing an input scene.
The algorithm requires a DL reasoner able to process a fuzzy ontology $\mathcal{O}$, which encodes the symbols related to types $\Gamma$ and relations $\mathbf{r}$.
We rely on a constant fuzziness value $a$, and on two thresholds to evaluate the classification confidence.
When the confidence is low, we learn and structure a new node of the memory graph from facts, which will be used to classify future input scenes.

At Lines~\ref{ln:econdingStart}--\ref{ln:econdingEnd}, the algorithm performs the encoding phase to assert the beliefs that represent the current scene $\epsilon_t$ in $\mathcal{O}$.
At Lines~\ref{ln:classifyStart}--\ref{ln:classifyEnd}, the memory graph $M$ is retrieved by querying all the subclasses of $\Phi$ from $\mathcal{O}$, and $M^\star_t$ is classified by selecting only the categories $\Phi_j\sqsubseteq\Phi$ having $\epsilon_t$ as their instance, \ie when
    ${p^{\class{\epsilon_t}{\Phi_j}}>0}.$
In addition, the similarity value $d^{\Phi_j}_{\epsilon_t}$ is computed and associated to each category of $M^\star_t$, which also encodes the membership value 
    $p^{\class{\epsilon_t}{\Phi_j}}$ 
computed by FuzzyDL.
 
Line~\ref{ln:ifStatament} assesses the confidence of the classification by considering the nodes that classify $\epsilon_t$ with the highest membership and similarity values.
Indeed, if 
    $p^{\class{\epsilon_t}{\Phi_j}}$ 
is lower than a threshold, then the scene beliefs do no enough respect cardinality restrictions, while, if 
    $d^{\Phi_j}_{\epsilon_t}$   
is too low, the classification would regards only a small portion of the scene $\epsilon_t$.
If one of these cases occurs, Line~\ref{ln:learn} is used to learn a new scene category $\Phi_t$ through $\mathcal{L}$.
At Line~\ref{ln:struct}, $\Phi_t$ is structured by $\mathcal{S}$ in the memory graph $M$ that is represented in the ontology $\mathcal{O}$.

Line~\ref{ln:reclassify} reclassifies the inputs with the new memory graph $M$.
Hence, the returned $M^\star_t$ always contains at least a node $\Phi_j$ and, if it has just been learned, it would have 
    $d^{\Phi_t}_\epsilon = 1$, and 
    $p^{\class{\epsilon_t}{\Phi_t}} = 1$ 
by construction.
Line~\ref{ln:clean} removes all the encoded beliefs in the ontology for processing future input scenes.

%\tikzexternaldisable
\begin{figure}[t]
    \footnotesize%
    \centering%
    \setlength{\fheight}{.17\textwidth}%
    \setlength{\fwidth}{.35\textwidth}%    
    \input{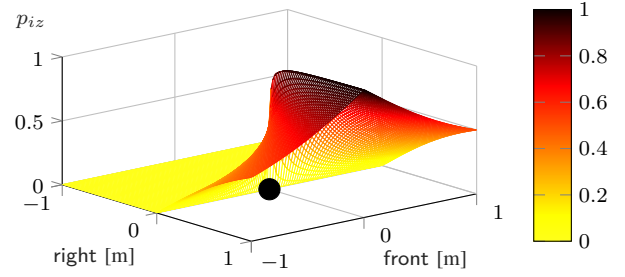}%    \input{./figure/right-kernel.tex}
    \caption{
        The fuzzy \emph{kernel} defining a \fuzzyprop{\gamma_x}{\gamma_y}{\onto{front}}{\,p_{iz}} axiom~\eqref{eq:in}.
        Given an \emph{element} of interest $\gamma_x$ located in (0,0), the kernel provides the degree $p_{iz}$ based on the relative position of the $\gamma_y$ element.
        The kernel is oriented based on a global reference frame to discriminate \onto{front} and \onto{right} relations. 
    }
    \label{fig:kernel}%
\end{figure}
%\tikzexternalenable

\begin{figure*}[ht]
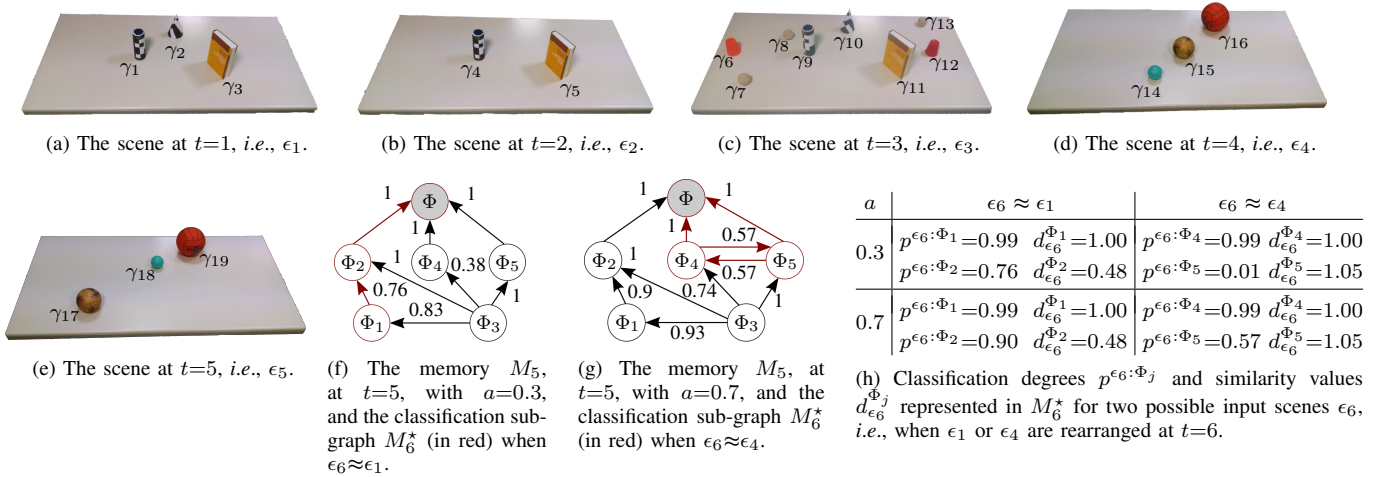
%
    \footnotesize
    \centering%
    \subfloat[The scene at $t{=}1$, \ie $\epsilon_1$.]{%
        \epstextikz{.23}{scenario1.eps_tex}%
        \label{fig:scenarios1}%
    }%
    ~%
    \subfloat[The scene at $t{=}2$, \ie $\epsilon_2$.]{%
        \epstextikz{.23}{scenario2.eps_tex}%
        \label{fig:scenarios2}%
    }%
    ~%
    \subfloat[The scene at $t{=}3$, \ie $\epsilon_3$.]{%
        \epstextikz{.23}{scenario3.eps_tex}%
        \label{fig:scenarios3}%
    }%
    ~%
    \subfloat[The scene at $t{=}4$, \ie $\epsilon_4$.]{%
        \epstextikz{.23}{scenario4.eps_tex}%
        \label{fig:scenarios4}%
    }%
    \\[-.5em]%
    \subfloat[The scene at $t{=}5$, \ie $\epsilon_5$.]{%
        \epstextikz{.23}{scenario5.eps_tex}%
        \label{fig:scenarios5}%
    }%
    %~%
    \subfloat[The memory $M_5$, at $t{=}5$, with $a{=}0.3$, and the classification sub-graph $M^\star_6$ (in red) when $\epsilon_6{\thickapprox}\epsilon_1$.]{%
        \epstextikz{.15}{memory30.eps_tex}%
        \label{fig:scenariosM1}%
    }%
    \hfill%
    \subfloat[The memory $M_5$, at $t{=}5$, with $a{=}0.7$, and the classification sub-graph $M^\star_6$ (in red) when $\epsilon_6{\thickapprox}\epsilon_4$.]{%
        \epstextikz{.17}{memory70.eps_tex}%
        \label{fig:scenariosM2}%
    }%
    \hfill%
    \subfloat[Classification degrees $p^{\class{\epsilon_6}{\Phi_j}}$ and similarity values $d^{\Phi_j}_{\epsilon_6}$ represented in ${M^\star_{6}}$ for two possible input scenes $\epsilon_6$, \ie when $\epsilon_1$ or $\epsilon_4$ are rearranged at $t{=}6$.]{%
        \footnotesize%
        \centering%
        \begin{tabular}[b]{@{}c@{~}|@{~}c@{~~}c@{~}|@{~}c@{~}c@{}}%
                  $a$
                & \multicolumn{2}{c@{~}|@{~}}{$\epsilon_6\thickapprox\epsilon_1$} 
                & \multicolumn{2}{c}{$\epsilon_6\thickapprox\epsilon_4$}                      \\[.15em]
            \hline&&&&\\[-0.85em]
            \multirow{2}{*}{$0.3$}
                & $p^{\class{\epsilon_6}{\Phi_1}}{=}0.99$  & $d^{\Phi_1}_{\epsilon_6}{=}1.00$       
                & $p^{\class{\epsilon_6}{\Phi_4}}{=}0.99$  & $d^{\Phi_4}_{\epsilon_6}{=}1.00$ \\[.15em]
                & $p^{\class{\epsilon_6}{\Phi_2}}{=}0.76$  & $d^{\Phi_2}_{\epsilon_6}{=}0.48$       
                & $p^{\class{\epsilon_6}{\Phi_5}}{=}0.01$  & $d^{\Phi_5}_{\epsilon_6}{=}1.05$ \\[.15em]
            \hline&&&&\\[-0.85em]
            \multirow{2}{*}{$0.7$} 
                & $p^{\class{\epsilon_6}{\Phi_1}}{=}0.99$  & $d^{\Phi_1}_{\epsilon_6}{=}1.00$       
                & $p^{\class{\epsilon_6}{\Phi_4}}{=}0.99$  & $d^{\Phi_4}_{\epsilon_6}{=}1.00$ \\[.15em]
                & $p^{\class{\epsilon_6}{\Phi_2}}{=}0.90$  & $d^{\Phi_2}_{\epsilon_6}{=}0.48$       
                & $p^{\class{\epsilon_6}{\Phi_5}}{=}0.57$  & $d^{\Phi_5}_{\epsilon_6}{=}1.05$ 
        \end{tabular}%
        \label{tab:classify}%
    }%
    \caption{
        Scenes perceived as noisy symbolic \emph{facts} over time, the representation that fuzzy SIT bootstrapped, and the classification of new scenes.
        Figures (a)--(c) shows a sequence of scenes $\epsilon_t$ that had been consequently learned as \emph{categories} $\Phi_t$.
        Scene categories were structured in the \emph{memory} graph $M_t$ for different \emph{fuzziness} values as shown in (f) $a{=}0.3$, and (g) $a{=}0.7$.
        The graph (f) also highlights the \emph{classification graph} $M^\star_6$ obtained when $\epsilon_1$ (a) was arranged again but with small differences (\ie $\epsilon_6{\thickapprox}\epsilon_1$) at $t{=}6$, while (g) highlights $M^\star_6$ when a new scene $\epsilon_6{\thickapprox}\epsilon_4$ was rearranged.
        Table (h) details the fuzzy classification degrees $p^{\class{\epsilon_6}{\Phi_j}}$ and similarity values $d^{\Phi_j}_{\epsilon_6}$ that was encoded in the $j$-th nodes of $M^\star_6$ with the two possible new scenes $\epsilon_6$.
        Since $\epsilon_6$ was always classified with high values in some $\Phi_j$, no category $\Phi_6$ have been learned at $t{=}6$, \ie $M_6$ had the same structure of $M_5$.
    }
    \label{fig:scenarios}%
\end{figure*}

\section{Fuzzy SIT Evaluation}
\label{sec:result}
\subsection{Evaluation Methodology}

% Scenario 
\noindent
We considered a scenario where the robot is a passive observer that acquires point clouds of tabletop scenes where objects are arranged by an user.
At each arranged scene, Algorithm~\ref{alg:algorithm} is computed and the outcomes are inspected.
We aim to evaluate the memory graphs that fuzzzy SIT bootstraps and the scene classification that it performs.
Section~\ref{sec:setup} describes the semantic of the scenes, which are defined with an input interface encoding scenes in terms of objects shapes and their spatial relations.

Section~\ref{sec:resMemory} focusses on the memory graph bootstrapped from a few demonstrations.
Since we want to compare SIT in the fuzzy and crisp domains, we replicated the scenes (shown in Figures~\ref{fig:scenarios1}--\ref{fig:scenarios5}) used in~\cite{sit} to evaluate the crisp implementation of SIT based on a similar input interface.
Since the crisp SIT is not robust to small input changes, in~\cite{dialogue_sit}, we proposed a dialogue-based supervision step to assure input correctness.
In this paper instead, we avoid the supervision step and we provide fuzzy SIT with input facts affected by perception noise.
Hence, we investigate the amount of vagueness that the fuzzy SIT can handle, while expecting to bootstrap memory graphs consistent with the crisp domain.

Section~\ref{sec:resClassify} presents another evaluation, which focusses on the robustness of fuzzy SIT by considering the classification distribution under nosy input facts.
We expect such a distribution to be smooth with respect to slightly different input facts, and that fuzzy SIT can consistently classify similar scenes. 

Finally, Section~\ref{sec:resComplexity} compares the complexity of fuzzy SIT among different sizes of the memory graph and input interface.

\subsection{Experimental Setup: an Input Interface Designing}
\label{sec:setup}

% Input interface rationale
\noindent
In our evaluations, we design an input interface concerning \emph{objects} as scene elements ($\gamma_x, \gamma_y$), their \emph{shape} as their types ($\Gamma_s, \Gamma_h$), and their mutual \emph{spatial relations} $\mathbf{r}_z$~\eqref{eq:in}.
In particular, we configure prior knowledge in the input interface to encompass 
    ${\Gamma\sqsubseteq\{\onto{SPHERE},}$ \onto{CONE}, \onto{CYLINDER}, \onto{PLANE}\}, and
    ${\mathbf{r}\sqsubseteq\{\onto{front},}$ \onto{right}, \onto{behind}, \onto{left}\}, 
which involve inverse and symmetric pairs of relations.

% Input Type degree
We perceive the shapes of objects from point cloud using RANSAC, and we identify a confidence value based on the ratio of the points of an object that match a given shape, which accuracy is reported in~\cite{buoncompagni2015software}.
Differently from the crisp domain, we do not assign to an object the shape with the highest confidence.
Instead, an object can be of different shapes simultaneously, with fuzzy degrees $p_{is}$ (or $p_{ih}$)~\eqref{eq:in} associated to the related confidences.

% Input Relation degree
The perception pipeline proposed in~\cite{buoncompagni2015software} was also used to perceive the centre of mass of each objects, which has been projected in the 2D plane associated with the workbench.
In the crisp SIT, spatial relations are identified through threshold-based rules involving pair of centre of mass~\cite{sit}.
Instead, in this paper we exploit fuzzy kernels as in~\cite{10.1007/978-3-030-03840-3_8}, and Figure~\ref{fig:kernel} shows the kernel to compute the degree of \onto{front} relations.
The kernel is centred on an object of interest $\gamma_x$ and defines a fuzzy degree $p_{iz}$~\eqref{eq:in} based on the relative position of another object $\gamma_y$.
The \onto{right} kernel is obtained with a rotation of $\pi/2$ with respect to a global reference frame placed on the workbench, and a similar rotation define \onto{behind} and \onto{left} as inverse and symmetric.

\subsection{Scenes Bootstrapping}
\label{sec:resMemory} 

% Result figure intro and fuzzy SIT consistency with crisp SIT
\noindent
When Algorithm~\ref{alg:algorithm} processed the $\epsilon_t$ scenes over time instants $t$ (as shown in Figures~\ref{fig:scenarios1}--\ref{fig:scenarios5}), it learned a scene category $\Phi_t$ for each $\epsilon_t$, and the final structured memory was the graph shown in Figure~\ref{fig:scenariosM1} when $a{=}0.3$, and Figure~\ref{fig:scenariosM2} when $a{=}0.7$.
Figure~\ref{fig:scenarios} shows that both memory graphs bootstrapped with fuzzy SIT have a logic representation consistent with the one obtained with the crisp SIT~\cite{sit} deployed in the same scenario.

% Input facts and encoding results
In particular, at $t{=}1$, we consider an empty memory graph $M_0$, and we encoded the beliefs related to $\epsilon_1$.
Since no classification could occur, the category $\Phi_1$ was structured as a node of the memory graph $M_1$.
The node only subsumed the root $\Phi$ with degree 1 due to the definition of $\mathcal{L}$~\eqref{eq:learn}.

At $t{=}2$, we perceived the input facts relate to the $\epsilon_2$ scene, where the $\gamma_4$ object was also classified as a \onto{CONE}, and the centres of mass of each object was nosily perceived. 
We report in~\eqref{eq:fact2} the input facts $F_2$ that we observed, whereas the relative beliefs encoded through $\mathcal{E}$~\eqref{eq:encoding} are shown in~\eqref{eq:bel2}.
\begin{align}
\label{eq:fact2}
    & \fuzzyclass{\gamma_4}{\onto{CONE}}{\,0.82},\;
      \fuzzyclass{\gamma_4}{\onto{CYLINDER}}{\,1},\;
      \fuzzyclass{\gamma_5}{\onto{PLANE}}{\,1},\notag\\
    & \fuzzyprop{\gamma_4}{\gamma_5}{\onto{right}}{\,0.16},\;
      \fuzzyprop{\gamma_4}{\gamma_5}{\onto{behind}}{\,0.84}.\notag\\
    & \fuzzyprop{\gamma_5}{\gamma_4}{\onto{left}}{\,0.16},\;
      \fuzzyprop{\gamma_5}{\gamma_4}{\onto{front}}{\,0.84}.\\[.5em]
\label{eq:bel2}
    & \prop{\epsilon_2}{0.17}{\reifyy{right}{CYLINDER}},\;
      \prop{\epsilon_2}{0.82}{\reifyy{behind}{CYLINDER}},\notag\\
    & \prop{\epsilon_2}{0.84}{\reifyy{front}{PLANE}},\;
~~~~\,\prop{\epsilon_2}{0.17}{\reifyy{left}{PLANE}},\notag\\
    & \prop{\epsilon_2}{0.17}{\reifyy{right}{CONE}},\;
~~~~~~\prop{\epsilon_2}{0.84}{\reifyy{behind}{CONE}}.
\end{align}
Since the cardinality of $\epsilon_2$ were not satisfied by the restrictions of $\Phi_1$, $\epsilon_2$ was not classified in any category of $M_1$.
Thus, a new category $\Phi_2$ was learned through $\mathcal{L}$~\eqref{eq:learn}, which defines a cardinality restriction ${^{a}\Omega(c_{zh})}$ for each $zh$-th combinations of beliefs shown in~\eqref{eq:bel2}, with related cardinalities $c_{zh}$.

%\tikzexternaldisable  
\begin{figure*}[t]
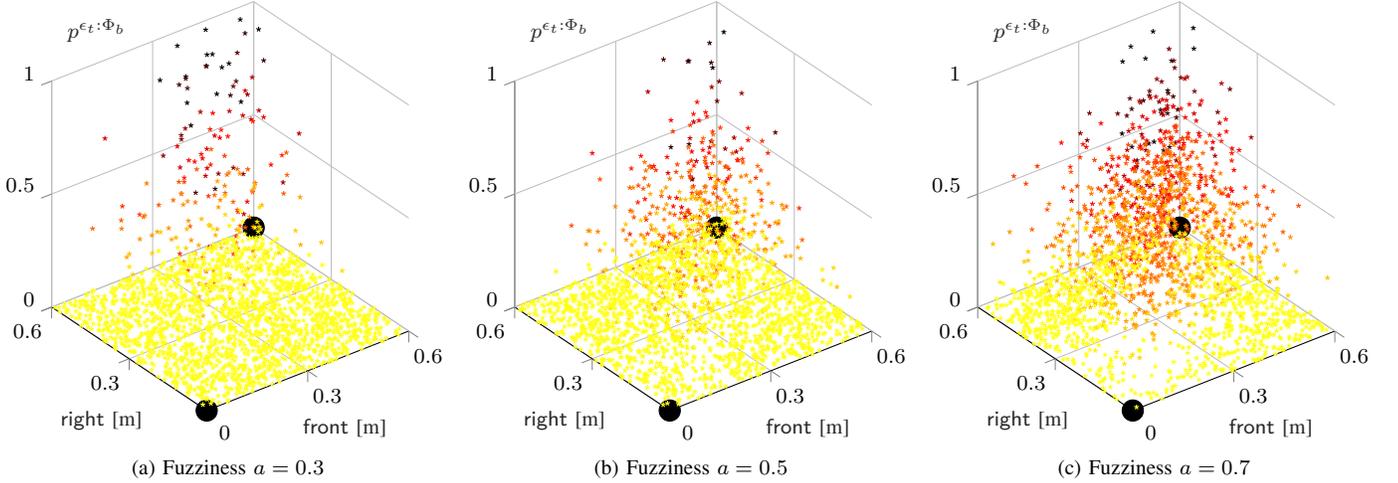
%
    \centering%
    \footnotesize%   
    \subfloat[Fuzziness $a=0.3$]{%
        \centering%
        \setlength{\fheight}{.3\textwidth}%
        \setlength{\fwidth}{.35\textwidth}%  
        \input{./figure/balanced03.tex}% \input{./figure/balanced03.tex}%
        \label{fig:balancedDistr03}%
    }%
    ~%
    \subfloat[Fuzziness $a=0.5$]{%
        \centering%
        \setlength{\fheight}{.3\textwidth}%
        \setlength{\fwidth}{.35\textwidth}%    
        \input{./figure/balanced05.tex}% \input{./figure/balanced05.tex}%
        \label{fig:balancedDistr05}%
    }%
    ~%
    \subfloat[Fuzziness $a=0.7$]{%
        \centering%
        \setlength{\fheight}{.3\textwidth}%
        \setlength{\fwidth}{.35\textwidth}%    
        \input{./figure/balanced07.tex}% \input{./figure/balanced07.tex}%
        \label{fig:balancedDistr07}%
    }%
    \caption{
        The fuzzy \emph{classification degree} distribution within the scenario introduced in Figure~\ref{fig:tennisGlass}.        
        Each plot shows the classification degree $p^{\class{\epsilon_t}{\Phi_b}}$ of 2500 scenes $\epsilon_t$ classified in the \emph{category} $\Phi_b$, which was learned from $\epsilon_b$ (Figure~\ref{fig:tennisGlassBalance}).
        In the plots, two spheres refer to the balls shown in Figure~\ref{fig:tennisGlass}, and the $p^{\class{\epsilon_t}{\Phi_b}}$ degree has been drawn relatively to the position that the glass had in each $\epsilon_t$.  
        The plots exploit the colormap in Figure~\ref{fig:kernel} and encompass different \emph{fuzziness} values. For small fuzziness values, only very similar scenes are classified (\ie $a{=}0$ reduces the classification to the crisp domain) while, for high fuzziness values, this condition is more relaxed.
        Note that the data shown in these plots is affected by perception noise in terms of object recognition and position.
    }%
    \label{fig:balancedDistr}%
\end{figure*}
%\tikzexternalenable

% structuring example
Then, the category $\Phi_2$ were structured by $\mathcal{S}$~\eqref{eq:struct} in the memory $M_1$, which got update in a new hierarchy $M_2$.
The latter graph represents $\epsilon_2$ as a sub-scene of $\epsilon_1$.
In other terms, FuzzyDL infers that the category $\Phi_1$ implies $\Phi_2$, \ie $\Phi_1\sqsubseteq\Phi_2$.
To better appreciate this implication, let us compare $\Phi_2$, which is deducible from~\eqref{eq:bel2}, and  $\Phi_1$~\eqref{eq:rest1}.
\begin{align}\label{eq:rest1}\small
    &\Phi_1 \triangleq \;\Phi \;\sqcap\; 
          \deff{\exists}{\reifyy{left}{CYLINDER}}{{^{a}\Omega(0.43)}} \,\sqcap\notag\\
        & \deff{\exists}{\reifyy{front}{CYLINDER}}{{^{a}\Omega(0.13)}}   \sqcap
          \deff{\exists}{\reifyy{right}{CYLINDER}}{{^{a}\Omega(0.35)}}   \sqcap\notag\\ %\,\sqcap\notag\\
        & \deff{\exists}{\reifyy{behind}{CYLINDER}}{{^{a}\Omega(0.9)}}   \sqcap          
          \deff{\exists}{\reifyy{front}{PLANE}}{{^{a}\Omega(0.78)}}    \,\sqcap\notag\\
        & \deff{\exists}{\reifyy{left}{PLANE}}{{^{a}\Omega(0.72)}}       \sqcap          
          \deff{\exists}{\reifyy{right}{CONE}}{{^{a}\Omega(0.94)}}     \,\sqcap\notag\\
        & \deff{\exists}{\reifyy{behind}{CONE}}{{^{a}\Omega(1)}}         \sqcap                          
          \deff{\exists}{\reifyy{left}{CONE}}{{^{a}\Omega(0.43)}}      \,\sqcap\notag\\
        & \deff{\exists}{\reifyy{front}{CONE}}{{^{a}\Omega(0.58)}}.                          
\end{align}
The comparison shows that $\Phi_2\not\sqsubseteq\Phi_1$ because $\Phi_1$ had more restriction that $\Phi_2$, which would never be satisfied. 
Instead, $\Phi_1\sqsubseteq\Phi_2$ because $\Phi_1$ had all $zh$-th combinations restricted by $\Phi_2$.
In particular, the \reifyy{behind}{CONE}, \reifyy{right}{CONE}, \reify{left}{PLANE}, \reify{behind}{CYLINDER}, \reify{right}{CYLINDER} cardinalities of $\Phi_1$ were always greater than in $\Phi_2$; hence, the related restrictions were satisfied (Figure~\ref{fig:restrictionYES}).
Furthermore, $\Phi_1$ had the cardinality related to the \reifyy{front}{PLANE} combination lower than $\Phi_2$, but its value was within the range defined by the fuzziness $a$, \ie the restriction was fuzzy satisfied (Figure~\ref{fig:restrictionFUZZY}).%
    \footnote{If the cardinality related to \reifyy{front}{PLANE} beliefs was not within the range specified by $a$ (Figure~\ref{fig:restrictionNO}), then FuzzyDL would have inferred that $\Phi_1\not\sqsubseteq\Phi_2$.}

Accordingly with Section~\ref{sec:struct}, fuzzy SIT used the minimum subsumption degree among $zh$-th combinations to identify the $p^{\Phi_1\sqsubseteq\Phi_2}$ degree, which was the one occurring in the \reifyy{front}{PLANE} combination.
Thus, the graph $M_2$ had the root $\Phi$ and two nodes: $\Phi_1$ and $\Phi_2$, which were connected with an edge starting from $\Phi_1$ and ending in $\Phi_2$ with a fuzzy degree $p^{\Phi_1\sqsubseteq\Phi_2}$ that depended on the fuzziness value, as also shown in Figures~\ref{fig:scenariosM1} and \ref{fig:scenariosM2}.
The implication $\Phi_1\Rightarrow\Phi_2$ represents that any scene classified as $\Phi_1$ would necessary also be classified as $\Phi_2$.
Therefore, the memory graph had more complex scenes toward its leafs, while simple configurations are closer to the root $\Phi$, which is consistent with the behaviour of crisp SIT.

At $t{=}3$, we encoded $\epsilon_3$, which was classified in $\Phi_1$ and $\Phi_2$ with an high fuzzy classification degree, but low similarity value.
Hence, the Line~\ref{ln:ifStatament} of Algorithm~\ref{alg:algorithm} leaded to bootstrap a new category.
We learned $\Phi_3$, which satisfied all the restrictions of $\Phi_2$, \ie ${p^{\Phi_3\sqsubseteq\Phi_2}=1}$, and fuzzy satisfied the restrictions of $\Phi_1$, \ie ${p^{\Phi_3\sqsubseteq\Phi_1}=0.76}$ with $a{=}0.3$ and ${p^{\Phi_3\sqsubseteq\Phi_1}=0.9}$ with $a{=}0.7$.
Consequentially, a new memory $M_3$ were structured as ${\Phi_1\sqsubseteq\Phi_2}$, ${\Phi_3\sqsubseteq\Phi_2}$, ${\Phi_3\sqsubseteq\Phi_1}$, with fuzzy subsumption degrees as shown in Figures~\ref{fig:scenariosM1} and \ref{fig:scenariosM2}.

At time $t{=}4$, we bootstrapped the category $\Phi_4$ in the memory $M_4$, and $\Phi_5$ was included at $t{=}5$ in $M_5$ (Figure~\ref{fig:scenarios}).
Since $\Phi_3$, $\Phi_4$ and $\Phi_5$ concerned three spheres placed in a similar configuration, $\Phi_4$ was implied by $\Phi_3$ in $M_4$, and the same also occurred for $\Phi_5$ in $M_5$.
Regardless from the used fuzziness value, $\Phi_1$ and $\Phi_2$ did neither imply $\Phi_4$ nor $\Phi_5$.

% threshould discussion
When we used crisp SIT in the same scenario, we could not discriminate the $\epsilon_4$ and $\epsilon_5$ scenes.
Indeed crisp SIT learned a category $\Phi_4$, and classified $\epsilon_5$ in $\Phi_4$~\cite{sit}, \ie the learning and structuring mechanism at $t{=}5$ were not performed.
In contrast, the fuzzy implementation could differentiate $\epsilon_4$ and $\epsilon_5$ by considering vague spatial relations rather then crisp thresholds.
In particular, the bootstrapping of $\Phi_5$ depends on the thresholds $\texttt{th}_m$ and $\texttt{th}_s$ used in Algorithm~\ref{alg:algorithm}.
Crisp SIT only requires $\texttt{th}_s$, which identifies the minimum similarity value $d_{\epsilon_t}^{\Phi_j}$~\eqref{eq:similarity} to discriminate sub-scenes so different to require a new scene bootstrapping.
Instead, fuzzy SIT also requires $\texttt{th}_m$, which concerns the minimum fuzzy classification degree $p^{\class{\epsilon_t}{\Phi_j}}$ under which we learn a new category (Section~\ref{sec:classify}).

% fuzziness discussion
With $a{=}0.3$, fuzzy SIT could not classify the scene $\epsilon_5$ since $M^\star_3$ was empty, \ie $p^{\class{\epsilon_5}{\Phi_4}}{=}0$, and we could not compute $d_{\epsilon_5}^{\Phi_4}$.
When $a{=}0.7$, the classification degree was $p^{\class{\epsilon_5}{\Phi_4}}{=}0.56$, and similarity $d_{\epsilon_5}^{\Phi_4}{=}0.95$.
Hence, fuzzy SIT learned $\Phi_5$ because either $\epsilon_5$ was not classified, or because the classification degree was lower than $\texttt{th}_m$.   
In addition, with low fuzziness values, $\Phi_4$ and $\Phi_5$ were considered as distinct categories, while with high fuzziness values, the cardinality restrictions $\Phi_4$ and $\Phi_5$ tends to overlap.
Categories with overlapping features represent qualitatively similar scenes, and they are denoted by nodes with mutual implications, as shown in Figure~\ref{fig:scenariosM2}.
When $a{=}0$, fuzzy SIT performed as the crisp SIT, which cannot have memory graphs with mutual implications since Algorithm~\ref{alg:algorithm} does not invoke the learning phase when a scene is classified.
Instead, when $a{=}1$, mutual implications are lucky to occur since fuzzy SIT relates all cardinalities of the same $zh$-th combination, but with low degree.
Generally, Figure~\ref{fig:scenariosM1} and \ref{fig:scenariosM2} shows that subsumption degrees increase when the fuzziness value increases because the relative sub-scenes features overlap more. %, even if they are subjected to perception noise.

% classification: similarity
Figure~\ref{fig:scenariosM1} and \ref{fig:scenariosM2} also highlight the classification graphs $M^\star_6$ we obtained through $\mathcal{C}$~\eqref{eq:classify} when we arranged again the scenes in Figure~\ref{fig:scenarios1} or \ref{fig:scenarios4} at $t{=}6$, \ie $\epsilon_6{\thickapprox}\epsilon_1$ or $\epsilon_6{\thickapprox}\epsilon_4$, respectively.
The graph $M^\star_6$ represented the classification in terms of fuzzy implications between some categories of $M_6$, and its nodes encode: the cardinality restrictions~\eqref{eq:learn}, the similarity value ${d_{\epsilon_t}^{\Phi_j}}$ and the classification degree ${p^{\class{\epsilon_t}{\Phi_j}}}$.
Table~\ref{tab:classify} shows the ${d_{\epsilon_t}^{\Phi_j}}$ and ${p^{\class{\epsilon_t}{\Phi_j}}}$ values encoded in the nodes of the classification graph for the two possible scenes arranged at $t{=}6$, and relative fuzziness values.
Consistently with the crisp implementation, the similarity values identified by fuzzy SIT was high in the leafs of $M^\star_t$ and they reduced toward the root $\Phi$.
As introduced in Section~\ref{sec:classify}, the similarity value of SIT is not linear with the number of objects in the scene, but with the number of relations.
Thus, similarity comparisons are more accurate when they involve scenes with a comparable number of objects, otherwise, the differences would rapidly increase.

% classification degree
At $t{=}7$, we arranged again the scene in Figure~\ref{fig:scenarios3}, \ie $\epsilon_7{\thickapprox}\epsilon_3$, and the classification graph $M^\star_7$ involved all the nodes shown in Figure~\ref{fig:scenariosM1} and \ref{fig:scenariosM2}, with the relative fuzziness value.
For instance, when we considered $a{=}0.7$, the classification degree and similarity values in $M^\star_7$ was
\begin{align}
    d_{\epsilon_6}^{\Phi_3}&{=}1,    & p^{\class{\epsilon_6}{\Phi_3}}&{=}0.97, & ~~~~~~
    d_{\epsilon_6}^{\Phi_5}&{=}0.2,  & p^{\class{\epsilon_6}{\Phi_5}}&{=}1,    \notag\\
    d_{\epsilon_6}^{\Phi_1}&{=}0.29, & p^{\class{\epsilon_6}{\Phi_1}}&{=}0.93, &
    d_{\epsilon_6}^{\Phi_4}&{=}0.19, & p^{\class{\epsilon_6}{\Phi_4}}&{=}0.74, \notag\\
    d_{\epsilon_6}^{\Phi_2}&{=}0.14, & p^{\class{\epsilon_6}{\Phi_2}}&{=}1.    
\end{align}
It is noteworthy that ${p^{\class{\epsilon_8}{\Phi_5}}>p^{\class{\epsilon_8}{\Phi_4}}}$.
This occurred because the scene in Figure~\ref{fig:scenarios3} has three spheres more similarly placed to the category $\Phi_5$ than $\Phi_4$.
The classification degree depends on the overlapping between encoded beliefs and cardinality restrictions induced by the fuzziness value $a$.
The next section also details this aspect and the differences of the similarity value in the crisp and fuzzy domains.

%\tikzexternaldisable  
\begin{figure}[t]
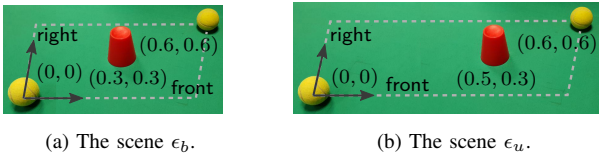

    \centering%
    \footnotesize%
    \subfloat[The scene $\epsilon_b$.]{%
        \epstextikz{.35}{TennisGlassBalanced.eps_tex}%
        \label{fig:tennisGlassBalance}
    }%
    \qquad%
    \subfloat[The scene $\epsilon_u$.]{%
        \epstextikz{.48}{TennisGlassUnbalanced.eps_tex}%
        \label{fig:tennisGlassUnbalance}
    }%
    \caption{%
        The scenario used to evaluate the classification distribution, where \emph{scenes} are obtained by moving the glass within the rectangle having the balls as vertexes.
        The scene learned to compute the classification shown in Figure~\ref{fig:balancedDistr} was (a), while (b) was used to obtain the distribution shown in Figure~\ref{fig:unbalancedDistr}.
    }%
    \label{fig:tennisGlass}%
\end{figure}
%\tikzexternalenable

% \tikzexternaldisable
\begin{figure}[t]
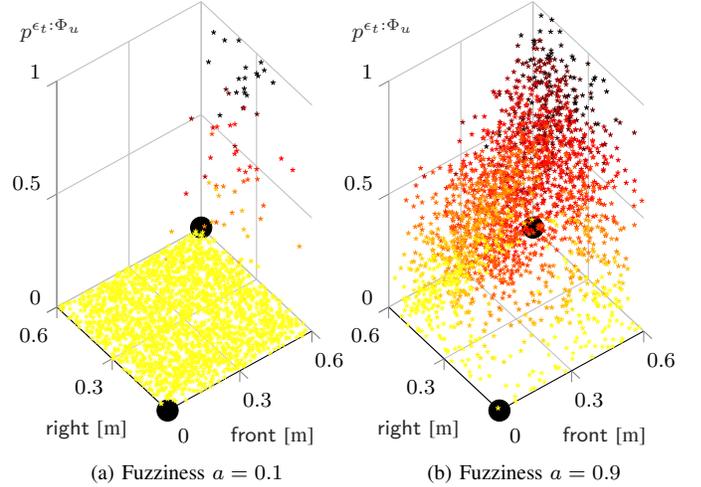

    \centering%
    \footnotesize%
    \subfloat[Fuzziness $a=0.1$]{%
        \hspace{-.5em}%
        \centering%
        \setlength{\fheight}{.3\textwidth}%
        \setlength{\fwidth}{.25\textwidth}%  
        \input{./figure/distrUmbalanced01.tex}% \input{./figure/distrUmbalanced01.tex}%
        \label{fig:classifyDistr1}
    }%
    \subfloat[Fuzziness $a=0.9$]{%
        \hspace{-1em}%
        \centering%
        \setlength{\fheight}{.3\textwidth}%
        \setlength{\fwidth}{.25\textwidth}%  
        \input{./figure/distrUmbalanced09.tex}% \input{./figure/distrUmbalanced09.tex}%
        \label{fig:classifyDistr2}
    }
    \caption{%
        The \emph{classification degree} distribution concerning $\Phi_u$, \ie a \emph{category} learned from $\epsilon_u$ (Figure~\ref{fig:tennisGlassUnbalance}).
        The plots are drawn as described in Figure~\ref{fig:balancedDistr}.
        In this case, we can note that the distribution is not centered as in Figure~~\ref{fig:balancedDistr}, but it changes relatively to the position of the glass in Figure~\ref{fig:tennisGlassUnbalance}.
    }%
    \label{fig:unbalancedDistr}%
\end{figure}
% \tikzexternalenable

\subsection{Classification Distributions}
\label{sec:resClassify}

% scenario
\noindent
Let us considered a scenario where two balls (\ie \onto{SPHERE}) delimit a rectangle containing a glass (\ie \onto{CYLINDER}), as shown in Figure~\ref{fig:tennisGlass}.
In this scenario, we bootstrapped memory graphs with one category, which is learned with the glass in a specific position, \ie as shown if Figure~\ref{fig:tennisGlassBalance} or \ref{fig:tennisGlassUnbalance}.
Then, we arranged new scenes by moving the glass, and we collected their fuzzy classification degrees and similarity values with respect to the previously learned category.
Figures~\ref{fig:balancedDistr} and \ref{fig:unbalancedDistr} show the classification distribution where, for each position assumed by the glass, we draw the classification degree of the respective scene.
For simplicity, Figures~\ref{fig:balancedDistr} and \ref{fig:unbalancedDistr} also show two spheres that represent the balls on the workbench.

To collect the data shown in Figure~\ref{fig:balancedDistr}, we encoded the scene $\epsilon_b$ as in Figure~\ref{fig:tennisGlassBalance}, and we learned the scene category $\Phi_b$
\begin{align}\label{eq:balanced}\small
    &\Phi_b \triangleq \;\Phi \;\sqcap\; 
          \deff{\exists}{\reifyy{left}{CYLINDER}}{{^{a}\Omega(0.74)}}   \,\sqcap\notag\\
        & \deff{\exists}{\reifyy{behind}{SPHERE}}{{^{a}\Omega(0.62)}}     \sqcap
          \deff{\exists}{\reifyy{right}{CYLINDER}}{{^{a}\Omega(0.67)}}    \sqcap\notag\\ %\,\sqcap\notag\\
        & \deff{\exists}{\reifyy{front}{CYLINDER}}{{^{a}\Omega(0.32)}}    \sqcap          
          \deff{\exists}{\reifyy{right}{SPHERE}}{{^{a}\Omega(0.94)}}    \,\sqcap\notag\\
        & \deff{\exists}{\reifyy{behind}{CYLINDER}}{{^{a}\Omega(0.66)}}   \sqcap                          
          \deff{\exists}{\reifyy{left}{SPHERE}}{{^{a}\Omega(0.78)}}     \,\sqcap\notag\\
        & \deff{\exists}{\reifyy{front}{SPHERE}}{{^{a}\Omega(0.72)}}.                          
\end{align}
Then, we collected 2500 scenes $\epsilon_t$ with the glass in different positions, and we evaluate their classification \class{\epsilon_t}{\Phi_b} with a fuzziness value $a{=}0.3$.
The same experiment was also performed with $a{=}0.5$ and $a{=}0.7$, and the relative  fuzzy classification degrees $p^{\class{\epsilon_t}{\Phi_b}}$ are shown in Figure~\ref{fig:balancedDistr}.

When $a$ was low, fuzzy SIT tended to its crisp implementation, and it could only classify the scenes having beliefs cardinality close to the one processed while learning $\Phi_b$.
Instead, when $a$ was high, fuzzy SIT classified more scenes having beliefs that differ from the one used while learning $\Phi_b$.
Indeed, with $a{=}0.3$, only 349 scenes out of $2500$ were classified with an average classification degree of $0.89$ and a standard deviation of $0.1$, whereas $1689$ scenes were classified with an average of $0.98$ and a standard deviation of $0.13$ when $a{=}0.7$.
For all the $a$ values, fuzzy SIT classified scenes accordingly with the learned category $\Phi_b$, and this is shown in Figures~\ref{fig:balancedDistr} since the higher classification degrees occurred when the glass was in the middle, \ie as shown in Figure~\ref{fig:tennisGlassBalance}.

Figure~\ref{fig:balancedDistr} shows that the classification distribution increased with the fuzziness value, and this gives an intuitive reason why scene overlapped more when $a{=}0.7$ then $a{=}0.3$ in the previous scenario (Section~\ref{sec:resMemory}).
Also, with a high $a$, each $zh$-th combinations of cardinality restrictions leaded to higher fuzzy degrees $p_{zh}$ when the cardinality $c_{zh}\in(k^-,k)$, as shown in Figure~\ref{fig:shoulder}.
Since the classification degree $p^{\class{\epsilon_t}{\Phi_b}}$ is given by the fuzzy conjunction of each $zh$-th combination~\eqref{eq:learn}, higher fuzziness values allowed for a smoother transition between the degrees of scenes that are (or not) classified.

We performed the same experiment also with the scene $\epsilon_u$ shown in Figure~\ref{fig:tennisGlassUnbalance}, which lead to the learned category 
\begin{align}\label{eq:unbalanced}\small
    &\Phi_u \triangleq \;\Phi \;\sqcap\; 
          \deff{\exists}{\reifyy{left}{CYLINDER}}{{^{a}\Omega(0.16)}}   \,\sqcap\notag\\
        & \deff{\exists}{\reifyy{behind}{SPHERE}}{{^{a}\Omega(1.01)}}     \sqcap
          \deff{\exists}{\reifyy{right}{CYLINDER}}{{^{a}\Omega(0.72)}}    \sqcap\notag\\ %\,\sqcap\notag\\
        & \deff{\exists}{\reifyy{front}{CYLINDER}}{{^{a}\Omega(0.85)}}    \sqcap          
          \deff{\exists}{\reifyy{right}{SPHERE}}{{^{a}\Omega(0.68)}}    \,\sqcap\notag\\
        & \deff{\exists}{\reifyy{behind}{CYLINDER}}{{^{a}\Omega(0.35)}}   \sqcap
          \deff{\exists}{\reifyy{left}{SPHERE}}{{^{a}\Omega(0.85)}}     \,\sqcap\notag\\
        & \deff{\exists}{\reifyy{front}{SPHERE}}{{^{a}\Omega(0.65)}}.                          
\end{align}
Figure~\ref{fig:unbalancedDistr} shows the classification distribution we observed with $a{=}0.1$ and $a{=}0.9$, and it highlights the effect of the fuzziness value, which was coherent with the one observed in Figure~\ref{fig:balancedDistr}.
Furthermore, the used input interface (described in Section~\ref{sec:setup}) were consistently able to represent the scenes characteristics. 
Indeed, also the distribution shown in Figure~\ref{fig:unbalancedDistr} show an higher classification degree in correspondence to the glass position related to $\Phi_u$ (Figure~\ref{fig:tennisGlassUnbalance}).

It is noteworthy that the plots in Figure~\ref{fig:balancedDistr} and \ref{fig:unbalancedDistr} show some outliers, which were due to a vague identification of the centre of mass and object type degrees.
Such a noise perturbed $p_{iz}$ and $p_{ih}$~\eqref{eq:in}, which affect $c_{zh}$ consequently~\eqref{eq:cardinalitySimplified}.
We designed the scenario shown in Figure~\ref{fig:tennisGlass} to stress the encoding faction~\eqref{eq:encoding} because scenes have the same $zh$-th combinations of beliefs, \eg~\eqref{eq:balanced} and~\eqref{eq:unbalanced}.
For maintaining this scenes characteristic, we used objects that were always perceived with the same shapes.

Figure~\ref{fig:balancedDistr} and \ref{fig:unbalancedDistr} shows that the fuzziness value $a$ could be tuned to make our algorithm robust to perception noise that generate reasonably small variances in the input facts.
Indeed, the 12500 scenes $\epsilon_t$ used to perform the experiments shown in Figures~\ref{fig:balancedDistr} and \ref{fig:unbalancedDistr} were subjected to noises affecting 
    ($i$)~the object centre of mass, with an average of $0.015$m and standard deviation of $0.038$m, and
    ($ii$)~the fuzzy degree associate to the object shapes, with an average of $0.093$ and a standard deviation of $0.195$. 
Figures~\ref{fig:balancedDistr} and \ref{fig:unbalancedDistr} confirmed that fuzzy SIT is able to classify scenes consistently with its crisp formulation also under this noisy input facts.

\begin{figure}[t]
    \centering%
    \footnotesize%
    \setlength{\fheight}{0.17\textwidth}%
    \setlength{\fwidth}{0.55\textwidth}%  
    \input{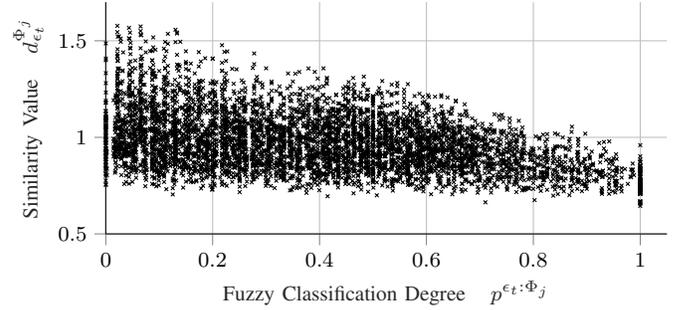}% \input{./figure/similarity.tex}%
    \caption{The \emph{similarity} distribution with respect to the fuzzy \emph{classification degree} among 5584 scenes arranged within the scenario shown in Figure~\ref{fig:tennisGlass}.}
    \label{fig:similExceed}%
\end{figure}

In the scenario shown in Figure~\ref{fig:tennisGlass}, the crisp SIT would assign a similarity value $d^{\Phi_j}_{\epsilon_t}{=}1$ to each scene classified in $\Phi_j$, \ie~\eqref{eq:balanced} or~\eqref{eq:unbalanced}.
Consistently, the fuzzy SIT computed an high similarity value, \ie with average $0.977$ and standard deviation of $0.147$.
However, as described in Section~\ref{sec:classify}, the fuzzy similarity might exceed 1, while the crisp similarity is bounded on $[0,1]$.
In particular, $d^{\Phi_j}_{\epsilon_t}{>}1$ occurred 2199 times out of 5584 classified scenes, which distribution with respect to the fuzzy classification degree $p^{\class{\epsilon_t}{\Phi_j}}$ as shown in Figure~\ref{fig:similExceed}.
The latter Figure shows a linear dependence between similarity and classification degree, and that a scene classified with a high $p^{\class{\epsilon_t}{\Phi_j}}$ tended not to have a similarity that exceeded $1$.

In the fuzzy implementation, the similarity might exceed 1 when a $zh$-th combination of restriction $\Omega(k)$ was fuzzy satisfied by the related cardinality of a scene $c_{zh}$, \ie ${k^{-}\leqslant c_{zh}\leqslant k}$ (Figure~\ref{fig:shoulder}).
In this case, it might be that $d^{\Phi_j}>d_{\epsilon_t}$~\eqref{eq:similarity} when the scene is classified, \ie $p^{\class{\epsilon_t}{\Phi_j}}>0$.
This would never occur in the crisp implementation since a scene is classified only if $c_{zh}\geqslant k$, \ie $d^{\Phi_j}>d_{\epsilon_t}$ always.
Therefore, $p^{\class{\epsilon_t}{\Phi_j}}>1$ occurred more frequently when 
    ($i$)~$p^{\class{\epsilon_t}{\Phi_j}}$ was low, \ie the restrictions were fuzzy satisfied, and 
    ($ii$)~the fuzziness $a$ was high, \ie $k^-$ was smaller.
Nevertheless, Figure~\ref{fig:similExceed} shows that the similarity $d^{\Phi_j}_{\epsilon_t}$ is bounded in a range, which upper-bound depends on the used input interface and fuzziness value.

% \tikzexternaldisable 
% \tikzexternalenable

%\tikzexternaldisable
\begin{figure*}[t]
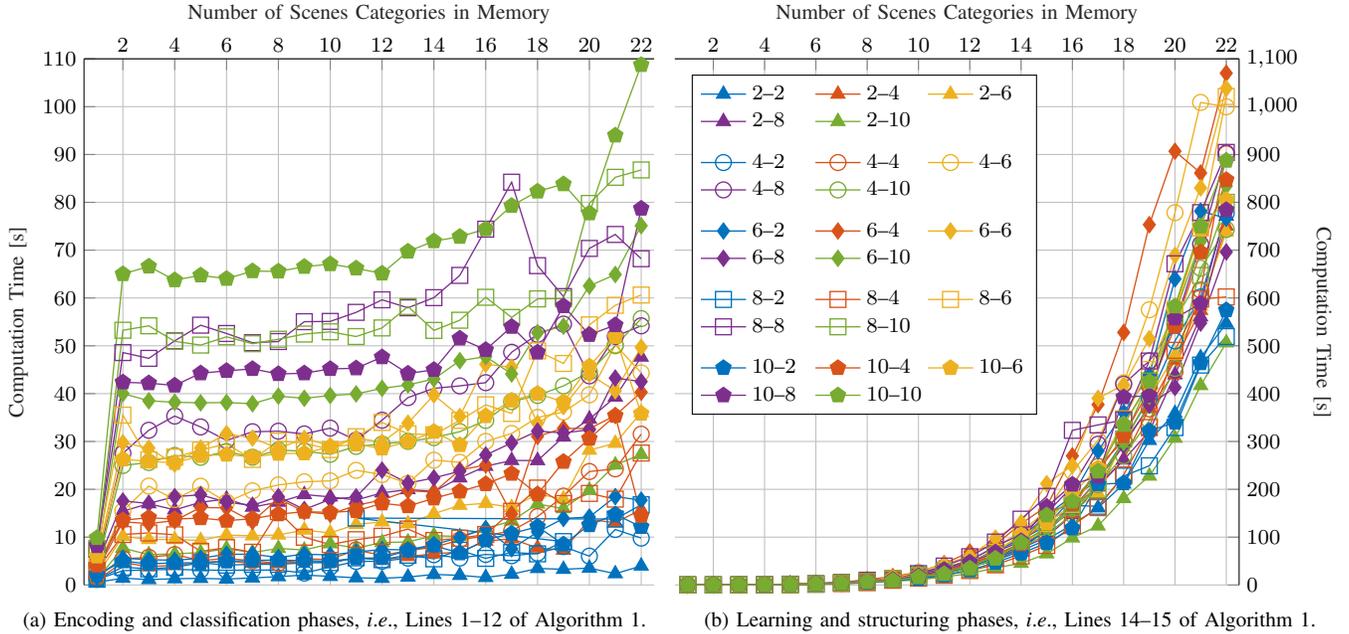

    \centering%
    \footnotesize%
    \setlength{\fheight}{7cm}%
    \setlength{\fwidth}{.43\textwidth}%  
    \subfloat[Encoding and classification phases, \ie Lines~\ref{ln:econdingStart}--\ref{ln:classifyEnd} of Algorithm~\ref{alg:algorithm}.]{%
        \centering%
        \input{./figure/recognitionScale.tex}% \input{./figure/recognitionScale.tex}%
        \label{fig:complexityRecognition}
    }%
    ~%\quad%
    \subfloat[Learning and structuring phases, \ie Lines~\ref{ln:learn}--\ref{ln:struct} of Algorithm~\ref{alg:algorithm}.]{%
        \centering% 
        \input{./figure/learningScale.tex}% \input{./figure/learningScale.tex}%
        \label{fig:complexityLearning}
    }%
    \caption{%
        The computation complexity against the number of scene categories, \ie nodes, in the memory graph $M$. The plots are shown for different number of element types $\Gamma$ and relationships $\mathbf{r}$ represented in the ontology as prior knowledge~\eqref{eq:in}, \ie $v$--$w$, respectively; as shown in the legend.
    }%
    \label{fig:complexity}%
\end{figure*}
%\tikzexternalenable

\subsection{Computation Complexity}
\label{sec:resComplexity}

% computation complexity
\noindent
Figure~\ref{fig:complexity} shows the computation complexity we observed with random scenes encoded through prior knowledge made by different combinations of $v$ and $w$ values.
In other words, we changed the amount of possible symbols representing the object types $\Gamma$ and relations $\mathbf{r}$ used to define the input interface, which spanned in ${\{2,4,\ldots,10\}}$ symbols.
We process $22$ scenes over time, and we incrementally learned the respective categories $\Phi_t$ in the memory.
In this scenario, we measured the time spent to learn and classify scenes in ontologies with a different number of axioms, \ie by modifying the number of $zh$-th combinations and the amount of categories in the memory.
Each point in the graph is given by the average of four measurements taken within Lines~\ref{ln:econdingStart}--\ref{ln:classifyEnd} of Algorithm~\ref{alg:algorithm} in Figure~\ref{fig:complexityRecognition}, and Lines~\ref{ln:learn}--\ref{ln:struct} in Figure~\ref{fig:complexityLearning}. 

We experienced an exponential increase of time when the number of axioms in the ontology increased, and this is due to FuzzyDL reasoner that is exponentially complex~\cite{f2016fuzzya,f2017Introduction}.
Noteworthy, DL-based reasoning has computation time with a high variance, which is also due to the open word assumption, and to the amount of knowledge that the reasoner can infer.
Thus, generalise reasoning complexity is difficult because the worst case scenario is not always representative of the typical behaviour of specific ontologies. 

Moreover, the exponential complexity introduced by the reasoner to perform the classification~\eqref{eq:classify} and structuring~\eqref{eq:struct} functions make the time required to perform the encoding~\eqref{eq:encoding} and learning~\eqref{eq:learn} functions negligible.
This occurs because~\eqref{eq:encoding} is linear with the number of facts, and~\eqref{eq:learn} with the amount of $zh$-th combinations.
Therefore, we can approximate the classification time \eqref{eq:classify} with Figure~\ref{fig:complexityRecognition}, and the structuring time \eqref{eq:struct} with Figure~\ref{fig:complexityLearning}.

Figure~\ref{fig:complexity} shows an exponential increase of complexity.
The trend of the classification complexity (Figure~\ref{fig:complexityRecognition}) is attenuated for input interfaces with a few relations $v$ and object types $w$, while is aggravated for big input interfaces.
However, the classification complexity was not largely affected by the number of categories in the memory.
In contrast, the structuring complexity (Figure~\ref{fig:complexityLearning}) was affected by memory categories, while big input interface had a relatively small contribution.

\section{Discussions}
\label{sec:discussions}

% Consistency
\noindent
Section~\ref{sec:result} shows that fuzzy SIT behaves consistently with its crisp counterpart since it preserves the properties concerning the encoding, learning, structuring and classification phases presented in Section~\ref{sec:fuzzySIT}.
In particular, fuzzy and crisp SIT share the same definition of the input interface, facts and encoded beliefs.
The learning function of fuzzy SIT defines scene categories through cardinality restriction as in the crisp domain.
The structuring function of fuzzy and crisp SIT arrange learned scene categories in a memory graph representing sub-scene through DL-based reasoning.
The classification provided by fuzzy and crisp SIT is a structure of learned categories retrieved through reasoning.
% Hence, fuzzy SIT extends crisp SIT, and the former becomes equivalent the latter when the fuzziness value $a{=}0$.

% What fuzzy SIT can do that crisp SIT cannot
Fuzzy SIT has also additional features encompassing the following statements.
    ($i$)~Elements in a scene can also be of different types at the same time, \eg \onto{CONE} and \onto{CYLINDER}.    
    ($ii$)~Similar scenes (\eg Figures~\ref{fig:scenarios4} and \ref{fig:scenarios5}) can be discriminated in the memory, and represented through implication cycles.%
        \footnote{In the crisp domain the memory graph is reduced to a tree-like hierarchy.}
    ($iii$)~The memory graph has edges weighed with the fuzzy subsumption degree, and
    ($iv$)~the classification includes fuzzy degrees. % , which were not computed by the crisp SIT.
While the crisp SIT classifies scenes with similarity values bounded in $[0,1]$, fuzzy SIT has the drawback to compute similarities values in a range with an upper bound depending on the used input interface and fuzziness value.

% Robustness
Section~\ref{sec:SITOverview} introduces the limitation of crsip SIT, which concern computation complexity and robustness to perception noises.
We shown that fuzzy SIT can overcome the robustness limitations by means of the fuzziness value $a$.
Fuzzy SIT is robust because allows a smooth transition between scene categories, while in the crisp domain there are discontinuities that lead to different scene representations even if the input facts have small differences.
Such a smooth transition is also due to the definition of the input interface~\eqref{eq:in} that extends crisp axioms with fuzzy degrees.
Hence, we could use fuzzy kernels (\eg Figure~\ref{fig:kernel}), instead of threshold-based crisp roles, and assign a confidence to the perceived object types, instead of choosing the best confident identification.

% Vagueness and intelligibility (human-computer interaction)
The fuzzy implementation can also deal with vague representation of scenes.
This is relevant for human-robot interaction, \eg when robots and users should not share formal knowledge.
Since SIT is based on a symbolic formalism, it bootstraps knowledge structures that can be intelligible to humans.
However, fuzzy SIT provides information that are less intuitive than in the crisp domain because the scenes beliefs counting mechanism is less natural. 
Both in the crisp and fuzzy domains, the intelligibility of the knowledge bootstrapped by SIT depends on the symbols involved in the input interface, and further works might be required for intuitively communicating them to a user.
As in the crisp implementation, also fuzzy SIT allows refining learned category through interaction.
Nevertheless, in contrast with our previous approach~\cite{dialogue_sit}, fuzzy SIT is robust enough to bootstrap a scene representation without asking humans to supervise the correctness of each input fact, which involves a not negligible effort from users.

% Incremental learning
As for the crisp SIT, the intelligibility of fuzzy SIT also benefits of the incremental bootstrapping of structured representations.
Indeed, having a classification of sub-scenes can help intelligibility if they have been annotated.
For instance, if the user makes the robot learning a category $\Phi_b$ from the scene in Figure~\ref{fig:tennisGlassBalance}, and if $\Phi_b$ is annotated as a \emph{balanced} scene, the classification degrees shown in Figure~\ref{fig:balancedDistr} would rank other balanced scenes.
Then, the user might bootstrap and annotate a set of relevant scene categories in the memory graph, and classify them in a more intelligible manner.
This is possible thanks to the incremental bootstrapping mechanism implemented by SIT, which allows to learn and structure new categories in the memory without the need to retune the algorithm based on the categories previously bootstrapped. 

% Generality and the importance of the input interface
The crisp and fuzzy SIT exploit a general-purpose input interface \eqref{eq:in}, which encodes some symbols as prior knowledge in the ontology to represent the elements in the scene, \ie their type and relations.
Thus it requires mechanisms to perceive fuzzy degrees related to such symbols from sensors.
In addition, since SIT bootstraps the memory and classify future scenes based combination of symbols provided in the input interface (\eg \reify{connected}{CABLE}), it can provide suitable outcomes only when the input interface allows representing the scene characteristics that are relevant for an application.

For instance, in the geometric scenarios presented in Section~\ref{sec:result}, we considered a symmetric input interface designed to avoid singularities, as discussed in Sections~\ref{sec:simplify} and \ref{sec:setup}.
It is noteworthy that te input interface of SIT is not limited to the geometric domain, \eg rooms characteristics might be used to bootstrap a topology for navigating robots, or temporal events might be used to segment a human-lead demonstrative task.
However, while crisp SIT only requires to ground symbols from sensor based on the input interface, fuzzy SIT also require to identify fuzzy degrees, which might involve challenging issues in some applications.
Also, it is always possible to extend the input interface by considering more combinations of beliefs, \eg \reify{in}{BIG}{CONTAINER}.
This approach would lead to bootstrapped representations that are more expressive, but it strongly affects computation performances. 

% Limitation: complexity
Since both crisp and fuzzy SIT exploit DL-based reasoner, which held most of the computation load, the two implementation has a comparable computation complexity.
In particular, SIT scales exponentially with the amount of axioms in the ontology, which depends on the number of symbols encoded in the input interface, and on the size of the memory graph.
Thus, reducing the computation load is a crucial issue for human-robot interaction.
A possibility is to segment a complex application into a set of smaller domains, each with a dedicated input interface, which encompass smaller ontologies orchestrated by a semantic network, \eg as proposed in~\cite{mon}.

Algorithm~\ref{alg:algorithm} is configured to learn each scene it could not classify. 
While this is suitable for specific applications (\eg when SIT can be enabled and disable based on user inputs), it might lead to bootstrapped memory graphs unnecessary big, \eg in a long-term interaction.
An approach to limit this issue, and decrease the computation load, is to assign a score to each learned category, which can be consolidated over time for maintaining in the memory only relevant categories, while the others would be forget.
We presented a framework for evaluating different heuristics to consolidate and forget scene categories in~\cite{sit_memory_ws,fuzzySITmemory2024}.
Another approach could be to periodically evaluate the memory graph in order to simplifying it by reasoning on the similarity among scene categories, which are represented as fuzzy subsumption degrees.

\section{Conclusion}

\noindent
We presented a fuzzy extension of the SIT algorithm, which incrementally bootstraps structured scenes representation encompassing sub-scenes and similar scenes in an OWL-DL ontology.
We shown that our extension is consistent with the crisp formulation of SIT, and it provides further features as well.
Differently from crisp SIT, fuzzy SIT is robust to perception noise, and it can process sensory data in realistic setups.
However, fuzzy SIT bootstraps scenes representations that are less intelligible than the one obtained with its crisp counterpart.

In particular, fuzzy SIT bootstraps a memory graph by learning from not annotated observations over time. 
When a new scene is observed, SIT classifies it and provide a graph representing scenes previously observed, that encodes similarity and implication measurements.
The memory bootstrapped by fuzzy SIT is based on a general-purpose input interface, which can be used to represent scenes with different semantics.

The main limitation of the SIT algorithm is the computation load held by the OWL-DL reasoner.
As further work we aim to address this limitation with the approaches proposed in Section~\ref{sec:discussions}.

\bibliographystyle{IEEEtran}
\bibliography{IEEEabrv,fuzzySITbib}
\enlargethispage{-5in}

\end{document}